\documentclass[11pt,a4paper]{article}
\usepackage{times,latexsym}
\usepackage{url}
\usepackage{tikz}
\usetikzlibrary{positioning,arrows.meta,fit,calc}
\usepackage{booktabs}
\usepackage{subcaption} 
\usepackage[T1]{fontenc}
\usepackage{booktabs}
\usepackage{multirow}
\usepackage{graphicx} 
\usepackage{makecell}
\usepackage{amsmath}
\usepackage{listings}
\usepackage{xcolor}

\newcommand{\blue}[1]{\textcolor{black}{#1}}

\usepackage[normalem]{ulem}
\newcommand{\name}{{WISE}}
\usepackage{subcaption}
\lstdefinestyle{smallpython}{
  language=Python,
  basicstyle=\ttfamily\fontsize{10.5pt}{12pt}\selectfont,
  keywordstyle=\bfseries,
  commentstyle=\itshape,
  showstringspaces=false,
  breaklines=true,
  breakatwhitespace=false,
  columns=fullflexible,
  keepspaces=true,
  frame=single,
  xleftmargin=0.5em,
  xrightmargin=0.5em,
  aboveskip=0.5em,
  belowskip=0.5em
}
\usepackage{amssymb}
\usepackage{wrapfig} 

\usepackage[acceptedWithA]{tacl2021v1}

\usepackage{xspace,mfirstuc,tabulary}

\usepackage{pifont}
\newcommand{\cmark}{\ding{51}}

\newif\iftaclinstructions
\taclinstructionsfalse 
\newcommand{\instr}[1]{} 

\iftaclinstructions
\renewcommand{\confidential}{}
\renewcommand{\anonsubtext}{(No author info supplied here, for consistency with
TACL-submission anonymization requirements)}
\fi

\iftaclpubformat 

\else

\fi

\title{Understanding Benchmark Language Under Weakened Formal Semantics}

\author{
Haoyang Chen$^{1}$\thanks{Equal contribution.} \and Kumiko Tanaka-Ishii$^{1}$\footnotemark[1]
\\
$^{1}$Department of Computer Science and Engineering
\\
School of Fundamental Science and Engineering
\\
Waseda University
\\
3-4-1 Okubo, Shinjuku-ku, Tokyo 169-8555, Japan
\\
\texttt{haoyangc0308@fuji.waseda.jp}
\\
\texttt{kumiko@waseda.jp}
}

\date{}

\begin{document}
\maketitle

\begin{abstract}
State-of-the-art NLP benchmarks require interpretation of
natural-language that specify conditions, procedures, and exceptions,
often relying on implicit assumptions and external knowledge.
Constructing complete semantic representations with proof-theoretic
guarantees is frequently impractical at scale, and purely text-based
reasoning offers limited means of inspection.  This paper asks how
much understanding of benchmark language can be achieved when formal
semantic guarantees are weakened.

We investigate this question by extracting \emph{computables}:
executable representations whose runtime behavior provides operational
evidence of semantic adequacy, including executability, execution
traces, and runtime failures.  We induce and iteratively refine
computables for benchmark instances using retrieval from external
knowledge.  Across mathematical reasoning, multi-step reasoning,
causal inference, and rule- and exception-heavy legal and biomedical
benchmarks, we find that the proposed approach consistently exceeds
text-only reasoning and one-shot code execution.  Beyond accuracy, our analyses show that these computables provide scalable, inspectable semantic evidence: they expose conditions and exceptions benchmark language forces into executable form, offering a practical bridge between proof-oriented semantics and purely textual reasoning.

\end{abstract}

\section{Introduction}

\blue{A long-standing goal of \textit{formal semantics} has been to make aspects of linguistic meaning explicit through structured representations that support systematic interpretation. From Frege's foundational work linking meaning to formal logic \citep{frege} to Montague's demonstration that fragments of natural language admit model-theoretic interpretation via typed lambda calculus \citep{montague1970ug,montague1973ptq}, this tradition has emphasized precise formal structures for representing semantic content. In some settings, this goal extends to proof-oriented environments such as Coq and Lean \citep{bertot2004coqart,demoura2015lean}, where representations are not only explicit, but also checked under rigorously defined rules.}

\blue{In this paper, we use \textit{strong} formal semantics to refer to this proof-oriented end of the spectrum: representations whose meaning is specified within a formal system and whose well-formedness or correctness is checked against that system. At the other end, free-form natural-language reasoning provides much weaker guarantees: intermediate commitments may remain implicit, difficult to inspect, and hard to verify step by step. Many contemporary NLP benchmarks fall between these two extremes. In domains such as law, medicine, mathematics, and causal reasoning, benchmark instances often involve procedural descriptions, background assumptions, exceptions, answer-choice structure, and external knowledge. These tasks often require more explicit structure than text-only reasoning provides, while remaining difficult to encode at scale in a fully proof-oriented formalism.}

\blue{This gap motivates what we call \textit{weakened} formal semantics. By this we do not mean a new proof-theoretic semantics with reduced axioms. Rather, we mean a weaker form of semantic evidence: intermediate representations that do not carry full formal guarantees, but still make interpretive commitments explicit enough to be inspected, executed, revised, and compared. The central question of this paper is therefore practical: when \textit{strong} formal guarantees are unavailable, what structured semantic evidence can still be extracted from benchmark language?}

\blue{In this work, we study benchmark language under such \textit{weakened} conditions through executable representations that we call \emph{computables}. Computables are programs induced from individual benchmark inputs whose runtime behavior---including successful execution, explicit checks, traces, and failures---provides operational evidence about what an interpretation commits to.} \blue{Our aim is not to propose a new proof-theoretic semantics, nor to equate executability with correctness. Rather, we ask whether executable, revisable intermediate representations can function as inspectable semantic evidence in settings where \textit{strong} formal guarantees are unavailable.}

\blue{Our approach is informed by recent work on mapping natural language to executable representations. Text-to-code models, including syntax- and AST-structured generation \citep{yin-neubig-2017-syntactic,rabinovich-etal-2017-abstract}, exploit program structure to enforce well-formedness, and recent reasoning frameworks execute generated programs to obtain answers \citep{chen2023programthoughtspromptingdisentangling,coc}. Notably, results in \citet{coc} suggest that while execution can improve reliability, the resulting code often plays a narrow role in the overall reasoning process and does not by itself yield a robust, systematic inference mechanism.}

\blue{We examine whether this limited inferential role persists when executables are treated not as one-shot tools for producing answers, but as revisable and inspectable intermediate objects. We instantiate this perspective with \name{} (Weakened Interpretation via Structured Execution) system, a retrieval-grounded iterative codification loop that refines executable programs using retrieved evidence. Across mathematical, multi-step reasoning, causal, legal, and biomedical benchmarks, we find that combining retrieval with iteration improves answer quality and makes intermediate semantic commitments more inspectable under \textit{weakened} formal guarantees. Beyond accuracy, we analyze induced programs through their traces, explicit checks, and failure modes to show which conditions each benchmark domain must externalize into runnable code.}



\begin{figure*}[t]
\centering
\includegraphics[width=\textwidth]{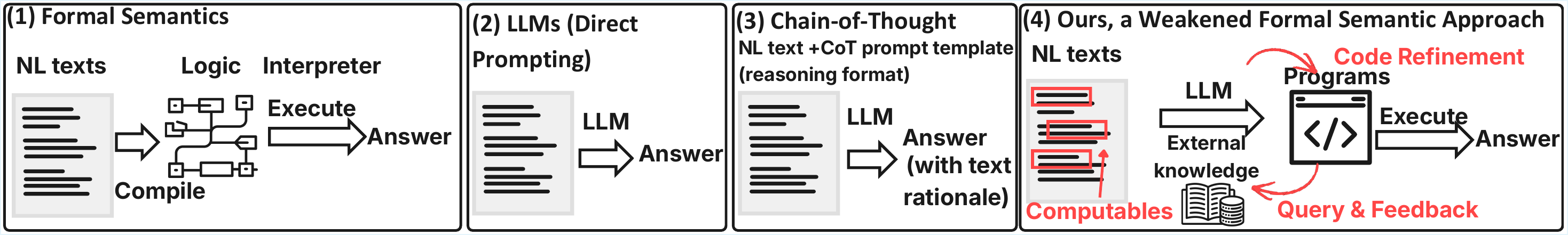}
\vspace{-3mm}
\caption{Four paradigms for mapping natural-language task descriptions to answers.}
\label{fig:paradigms}
\vspace{-3mm}
\end{figure*}

\section{Related Work}
\label{sec:relate}
\subsection{\blue{Formal Semantics and Structured Intermediate Representations}}

Formal semantics has classically treated meaning as compositional and representable through precisely defined calculi \citep{frege,montague1970ug,montague1973ptq,heim1998semantics,Winter2016Elements,Barwise1981Generalized}. Strong versions of this tradition emphasize representations that support mechanically defined interpretation or checking, as in proof-oriented environments such as Coq and Lean \citep{bertot2004coqart,demoura2015lean}. These traditions provide an important point of contrast for our work: benchmark language often calls for explicit interpretation, but in practice it is difficult to supply fully specified, proof-oriented representations at benchmark scale.

\blue{A closer methodological point of comparison is work on structured intermediate representations for language understanding. In particular, Discourse Representation Theory (DRT) builds explicit representations for discourse in which entities, references, and conditions are written into a structured form, rather than left implicit in text. Computational work on DRT further connects such representations to executable inference procedures \citep{kamp1993discourse,blackburn2005representation,bos2008wide}. Unlike these traditions, our computables are not intended as linguistically canonical semantic representations, nor do they inherit the same model-theoretic commitments. The connection is instead methodological: both lines of work treat intermediate representations as objects that can be inspected, manipulated, and linked to downstream reasoning procedures.}

Our use of weakened formal semantics is therefore practical rather than foundational. We do not claim full formal adequacy. Instead, we ask whether, when stronger guarantees are unavailable, executable intermediate representations can still support constrained, inspectable, and revisable interpretation.

\subsection{Executable Reasoning and Code as Semantic Representation}
\label{sec:exec_reasoning}
A parallel line of work in NLP maps natural-language inputs to
executable or formal representations.  Classic neural code generation
emphasizes syntactic well-formedness by making structure explicit, for
example through grammar- and AST-structured decoding 
\citep{yin-neubig-2017-syntactic,rabinovich-etal-2017-abstract},
thereby establishing code as a viable intermediate semantic
representation linking language and computation.

More recent reasoning frameworks use execution as a check on intermediate steps.
Chain-of-Thought and self-consistency expose intermediate reasoning but
remain primarily textual
\citep{CoT,wang2023selfconsistencyimproveschainthought}, while their
limitations as semantic evidence have been documented
\citep{zheng2025cursecotlimitationschainofthought}.
Search-based inference explores structured spaces of candidate reasoning
trajectories
\citep{10.5555/3666122.3666639,10.1609/aaai.v38i16.29720}.
Complementarily, several approaches express intermediate reasoning in executable code, including PAL, Program-of-Thoughts, and Chain-of-Code \citep{10.5555/3618408.3618843,chen2023programthoughtspromptingdisentangling,coc}. \blue{Related work has also explored logic-programming intermediates in narrower domains: \citet{yang2024prolog} generate Prolog programs for arithmetic reasoning and use predicate permutation as data augmentation, showing that executable symbolic representations can outperform text-only reasoning in math-specific settings.} Across these lines, execution is typically used to produce or check an answer for the current instance. In particular, results in \citet{coc} suggest that the benefit of code-mediated reasoning is uneven across task types, with weaker performance on causal-inference-style problems, indicating that execution alone does not guarantee strong inference in settings that require relational or counterfactual reasoning.

Our work aligns with this literature in treating executability as a
useful constraint, but differs in how executable representations are
used.
Where most prior approaches employ generated code as a one-shot
instrument for producing answers, we treat executable artifacts as
inspectable semantic objects, whose structure and runtime behavior can
be analyzed, compared, and incrementally revised under weakened formal
guarantees.

\subsection{Enhancing Reasoning with Retrieval and Iteration}

Retrieval augmentation grounds reasoning in external evidence and has
been shown to improve language modeling and few-shot performance
\citep{RAG,borgeaud2022improvinglanguagemodelsretrieving,
izacard2022atlasfewshotlearningretrieval}.
Interleaving retrieval with intermediate reasoning steps further
benefits multi-hop and knowledge-intensive benchmarks
\citep{trivedi2023interleavingretrievalchainofthoughtreasoning,
wang2024ratretrievalaugmentedthoughts}.
In parallel, work on code generation and maintenance demonstrates that
iterative feedback and revision improve reliability, including
retrieval-augmented code generation and program revision
\citep{parvez2021retrievalaugmentedcodegeneration,
wang-etal-2025-coderag,xia,jimenez2024swebenchlanguagemodelsresolve,
jiang,10.1007/s10515-024-00451-y}.
Our work extends single-pass codification to obtain executable
representation by integrating retrieval to supply missing background
information and incremental revision to iteratively improve the
induced executable representations.

\begin{figure*}[t]
\begin{center}
\centerline{\includegraphics[width=\linewidth]{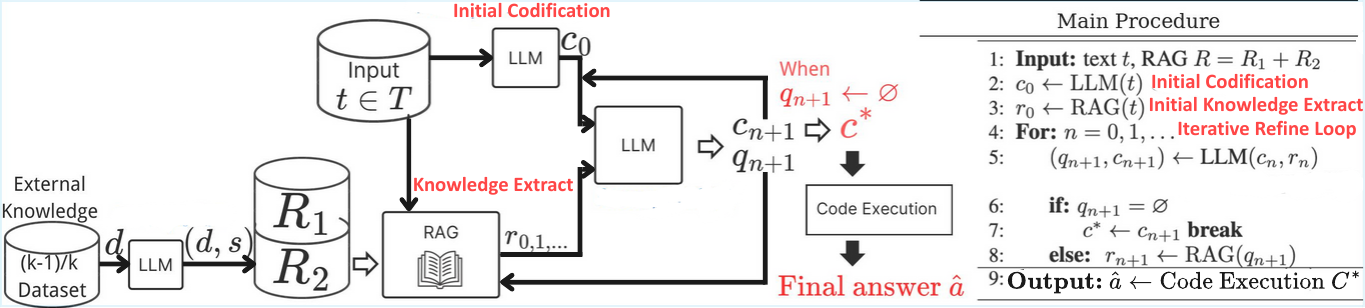}}
\vspace{-3mm}
\caption{Flowchart (left) and pseudocode (right) of our system.}
\label{flowchart}
\end{center}
\vspace{-1cm}
\end{figure*}

\begin{table}[t]
\centering
\small
\setlength{\tabcolsep}{2.4pt}
\renewcommand{\arraystretch}{1.12}
\caption{Comparison of paradigms in Figure~\ref{fig:paradigms}.}
\vspace{-3mm}
\label{tab:paradigm-tradeoff}
\resizebox{\columnwidth}{!}{%
\begin{tabular}{lcccc}
\toprule
\textbf{Dimension} &
\textbf{Formal} & 
\textbf{LLM} & 
\textbf{CoT} &
\textbf{Ours} \\
&
\textbf{Semantics} & 
& 
&
\\
\midrule
Formal guarantees  & High    & None   & None   & Partial \\
Accountability     & High    & Low    & Medium & High \\
Coverage           & Low     & High   & High   & High \\
Inspectability      & High    & Low    & Medium & High \\
\bottomrule
\end{tabular}}
\vspace{-3mm}
\end{table}

\section{Computational Paradigms for Language Interpretation}
\label{sec:paradigms}

Figure~\ref{fig:paradigms} illustrates computational paradigms for
mapping natural-language inputs to answers.
Table~\ref{tab:paradigm-tradeoff} provides a qualitative comparison of
these paradigms.  They differ primarily in the form of intermediate
representation and, as summarized in the first column of
Table~\ref{tab:paradigm-tradeoff}, exhibit different degrees of formal
guarantees, accountability, coverage across domains, and
inspectability.

At one end of the spectrum (left (1)), classical formal semantics maps
language to explicit formal representations whose interpretation is
governed by an interpreter or proof system
\citep{frege,montague1970ug,montague1973ptq,heim1998semantics,
  Winter2016Elements}.  At the other end (2), LLM-only prompting
produces answers directly through latent inference, offering broad
coverage but limited transparency.  Between these extremes lie
approaches that introduce intermediate representations with varying
degrees of structure.  Chain-of-Thought exposes intermediate reasoning
steps in text \citep{CoT,wang2023selfconsistencyimproveschainthought}
(3).  Under \emph{weakened formal semantics} (4), interpretation proceeds
through formal but incomplete executable representations whose
adequacy is assessed operationally rather than by proof.  Recent
executable approaches enabled by LLMs (Section~\ref{sec:exec_reasoning}) demonstrate the
usefulness of code for reasoning, but typically treat generated
programs as one-shot instruments for obtaining answers.

Our paradigm belongs to this family and instantiates weakened formal
semantics by treating executable programs as persistent, inspectable
objects that can be iteratively refined and analyzed as semantic
evidence.  Rather than advancing a new semantic theory, our aim is to
investigate how far language interpretation can proceed when formal
guarantees are weakened but not abandoned.  Executable representations
are treated as meaning-bearing artifacts evaluated through operational
evidence obtained by execution.  A key distinguishing feature is the
integration of retrieval augmentation and iterative refinement into
the interpretation process; to our knowledge, among benchmark-focused workflows that induce executable programs, \name{} is one of the first to tightly couple retrieval with execution-driven iteration to stabilize and refine the induced computables.

Throughout the paper, we use \emph{formal} to denote representation in a well-defined formal language, including both logical calculi and programming languages.
We refer to executable representations induced from natural-language inputs as \emph{computables}.
We call our approach \emph{weakened formal semantics}: instead of requiring proof-theoretic validity or complete logical coverage, we treat a computable as an admissible semantic representation as long as it is \emph{runnable} and its behavior can be inspected and revised.
Execution of a computable yields \emph{operational evidence}, such as successful runs, execution traces, and runtime failures, which makes the interpretation accountable in the limited sense that missing conditions and incorrect commitments can be localized and corrected.

\section{Implementation}
\label{sec:method}

This section describes \name{}, our executable system for instantiating \emph{weakened formal semantics}.
Conceptually, weakened formal semantics relaxes proof-theoretic guarantees and instead treats a representation as admissible when it can be executed, inspected, and revised using operational feedback.
In our implementation, each natural-language input is compiled into an executable Python program (a \emph{computable}) and iteratively revised with retrieved evidence and execution feedback; successful runs, traces, and runtime failures make missing conditions and incorrect commitments localizable and thus support targeted refinement (Section~\ref{sec:computables}).

\blue{Unless otherwise noted, all experiments
  in this paper use ChatGPT-5 with temperature $0.2$.
  We use the same
  model for all benchmarks to avoid confounding workflow effects with
  model variation.}

\subsection{Problem Setting}

Each benchmark instance includes a natural-language input $t \in
\mathcal{T}$ paired with a target answer $a \in \mathcal{A}$, forming
a dataset of instances $(t,a) \in \textsc{Dataset}$.  Given an input
$t$, the system induces a Python program $c$, a computable, whose
execution produces a predicted answer $\hat{a}$.
Depending on the benchmark, the answer space $\mathcal{A}$ may be
numeric, categorical, or structured (e.g., proofs), and evaluation is
performed against the gold standard answer $a$ provided in the
\textsc{Dataset}.

\begin{figure*}[t]
\centering

\lstset{
  language=Python,
  basicstyle=\ttfamily\scriptsize,
  columns=fixed,
  breaklines=true,
  frame=single,
  framerule=0.8pt,
  rulecolor=\color{black},
  escapeinside={(*!}{!*)},
  moredelim=[is][\color{red}]{(*@}{@*)},
  aboveskip=4pt, belowskip=4pt,
  showspaces=false,
  showstringspaces=false
}

\begin{minipage}[t]{0.43\textwidth}
\centering
\textbf{(a) $c_0$ (initial) --- Predicted: (B)}\\[-1mm]
\begin{lstlisting}
def solve(t, choices):
    s = t.lower()
    pregnant = ("pregnan" in s) or ("weeks pregnant" in s)
    seizure = ("seizure" in s)

    # Under-specified: detects key cues
    # but lacks an explicit action rule.
    if pregnant and seizure:
        return "B"  # default: magnesium + monitoring

    return fallback(choices)
\end{lstlisting}
\end{minipage}\hfill
\begin{minipage}[t]{0.55\textwidth}
\centering
\textbf{(b) $c^\star$ (refined, $n{=}1$) --- Predicted: (C)}\\[-1mm]
\begin{lstlisting}
def solve(t, choices):
    s = t.lower()
    pregnant = ("pregnan" in s) or ("weeks pregnant" in s)
    seizure  = ("seizure" in s)
    severe_bp = ("163/101" in s) or ("mmhg" in s)
(*@# retrieved rule: pregnancy + severe BP + seizure @*)
(*@# magnesium for seizure; proceed to delivery.@*)
    if pregnant and seizure and severe_bp:
(*@        c = choices["C"].lower()@*)
(*@        if ("magnesium" in c) and ("cesarean" in c):@*)
(*@            return "C"@*)
        return "B"

    return fallback(choices)
\end{lstlisting}
\end{minipage}
\vspace{-5mm}
\caption{One-step refinement on a USMLE multiple-choice medical QA item. Red marks code added after retrieval. Executing $c_0$ predicts option (B); executing $c^\star$ predicts option (C), matching the benchmark's reference label.}
\vspace{-3mm}
\label{fig:usmle-example}
\end{figure*}

\subsection{Overview of Our System}

Figure~\ref{flowchart} illustrates the overall system, with the left
panel showing the data flow and the right panel summarizing the
procedure in pseudocode.
Given a natural-language input $t$, the system induces an executable
representation through an \textbf{iterative} compilation process guided
by retrieval from external knowledge.

\paragraph{Compilation (with \textbf{iteration}).}
The process begins with \emph{initial codification} (upper middle
Fig.~\ref{flowchart}), in which an LLM translates the input $t$ into
an initial executable program $c_0$. An initial retrieval step
extracts an evidence $r_0$ from the external knowledge source $R$
(Section~\ref{sec:rag}; Fig.~\ref{flowchart}, bottom-left), using $t$ or a query derived from it.
Compilation then proceeds through \textbf{iteration} (middle).  At each iteration
$n \ge 0$, the model is conditioned on the current pair $(c_n, r_n)$
and produces (i) a revised program $c_{n+1}$ and (ii) a retrieval
query $q_{n+1}$ that targets missing information suggested by the
current program.  The query is used to retrieve additional evidence
$r_{n+1} \leftarrow \textsc{Retrieve}(q_{n+1})$, which is incorporated
into the next revision step.  In this way, retrieval and
iteration are integral to compilation, enabling progressive
refinement of the executable representation.  Compilation terminates
when the model emits no further retrieval queries
($q_{n+1}=\varnothing$) or when a maximum number of rounds is reached,
yielding a final executable $c^\star$ (right in Fig.~\ref{flowchart}).

\paragraph{Execution.}
\label{sec:execution}

After compilation terminates, we execute the final code $c^\star$ to obtain the prediction $\hat{a}$.
If execution raises a runtime exception, we record its type and message and count it toward the \textbf{Exception Ratio} in Table~\ref{tab:main_combined}.
Beyond producing an answer, execution provides operational evidence—success/failure and traces—about whether the induced procedure conforms to the benchmark protocol and where it remains underspecified.
These summary statistics of execution outcomes support the analyses in Section~\ref{sec:computables}.

\subsection{Retrieval Sources}
\label{sec:rag}
The system retrieves external evidence from a knowledge base $R$ to
support iterative compilation.
In our implementation, $R$ comprises two sources (Fig.~\ref{flowchart} bottom-left).
The first, $R_1$, is a domain-specific corpus associated with each
benchmark, such as statutes for legal benchmarks, biomedical abstracts
for medical QA, or task-provided information when available.
The second, $R_2$, is a pool of solved exemplars derived from the
benchmark dataset itself and constructed in a leakage-safe manner.

To avoid evaluation leakage that the model retrieves the same instance from $R_2$, we construct $R_2$ with $k$-fold
partitioning.
For each evaluation fold, the retrieval pool excludes all instances in the held-out fold and is built only from the remaining $(k-1)/k$ portion of \textsc{Dataset}.
Each example $d$ in this pool is transformed into a short Python
snippet $s=\textsc{LLM}(d)$ using a snippet-generation prompt, and the
resulting snippets are stored for retrieval (Fig.~\ref{flowchart}, bottom-left).
This yields a repository of \emph{program} exemplars that provide
structural guidance during compilation without copying solutions to
evaluation instances.

\subsection{Illustrative Example (USMLE: emergency management)}
\label{sec:illustrative-example}

{ This section provides an example illustrating how a task is
  solved using our system. More examples that cannot be solved
  directly by GPT-5 are provided in Appendix B.}

\noindent\textbf{Example (USMLE benchmark; the labeled correct option is \textbf{C}).}
A 24-year-old woman at 36 weeks pregnant presents with headache and abdominal pain.
Her blood pressure is 163/101 mmHg.
Before the physical exam, she develops a generalized seizure lasting 60 seconds that resolves spontaneously.
What is the next step?
Choices: (A) Diazepam, magnesium, and continuous monitoring; (B) Magnesium and continuous monitoring;
(C) Magnesium and cesarean section; (D) Nifedipine and cesarean section.

Figure~\ref{fig:usmle-example} illustrates retrieval-driven refinement that converts an initial cue-based program into an auditable decision procedure.
Executing $c_0$ predicts option (B): the program detects ``pregnant'' and ``seizure'' but lacks an explicit condition linking the described severity to the required action, so it falls back to monitoring.
After retrieval (content summarized in red in $c^\star$), $c^\star$ makes the missing condition explicit (\texttt{severe\_bp}) and encodes the retrieved management rule as executable branching.
It then grounds the rule to the answer options by selecting the unique choice whose text contains both required actions (``magnesium'' and ``cesarean''), yielding option (C), which matches the benchmark label.

\section{Experimental Settings}
\label{sec:exp}

We evaluate \name{} across a diverse set of benchmarks.
Section~\ref{sec:mainresults} reports quantitative benchmark results focused on final answer accuracy.
The primary evaluation, however, is presented in Section~\ref{sec:computables}, where we investigate how far a weakened formal semantics approach can be pushed by analyzing the behavior of computables.

\blue{The main body reports the full benchmark analysis using GPT-5 as the primary model, since this allows a consistent investigation of the proposed framework across all benchmark families and all downstream analyses of induced computables.
{Full settings and results are reported in {Appendix A,C}.}
To address generalizability beyond a single model family, we additionally include cross-model experiments, reported in {Appendix D}. }


\subsection{Benchmarks}
\label{sec:benchmarks}

We evaluate our approach on a diverse collection of benchmarks
designed to probe different forms of structured interpretation.
These include
(i) mathematical reasoning and proof-like problems\footnote{GSM8K~\cite{gsm8k} and ProofNet~\cite{proofnet}.},
(ii) causal reasoning tasks\footnote{IFQA~\cite{ifqa} and Corr2Cause~\cite{jin2024largelanguagemodelsinfer}.},
(iii) broad multi-step reasoning suites drawn from Big-Bench families\footnote{Big-Bench-Hard (BBH)~\cite{suz2022} and Big-Bench-Extra-Hard (BBEH)~\cite{kazemi2025bigbenchextrahard}.},
and (iv) high-stakes, rule- and exception-heavy domains such as legal judgment\footnote{CAIL~\cite{CAIL}, ECHR~\cite{ECHR}, and CAP~\cite{CAP}.}
and biomedical or health reasoning\footnote{Health-Claim~\cite{health}, PubMedQA~\cite{jin2019pubmedqa}, and BioASQ-6B~\cite{bio}.}. Together, these
benchmarks span settings with varying degrees of procedural structure,
implicit assumptions, and external knowledge requirements.

Table~\ref{tab:main_combined} summarizes each dataset and reports two classes of measurements.
The \textbf{Knowledge} column (2nd column) indicates the primary source used for $R_1$ retrieval in our system, and \textbf{Accuracy} (columns 3--10) compares final task performance across prompting baselines, ablations, and our iterative codification method.

Beyond accuracy, Table~\ref{tab:main_combined} reports three \text{summary statistics of execution outcomes} (right-most columns).
$N$ (Avg/Max) is the number of refinement rounds used by our iterative codification.
\text{Executability} is the fraction of instances whose induced program runs to completion and returns an answer.
\text{Exception Ratio} is the fraction of instances that raise any runtime error during execution.

\begin{table}[t]
\centering
\caption{Baselines are constructed as component-wise ablations over retrieval, executability, and iteration.
\textbf{Output Form} indicates whether the final answer is obtained from text generation (Text) or program execution (Program).}
\label{tab:baseline}
\footnotesize
\setlength{\tabcolsep}{3pt}
\begin{tabular}{lcccc}
\toprule
\textbf{Method} & \textbf{Retrieval} & \textbf{Exec} & \textbf{Iteration} & \textbf{Output Form} \\
\midrule
LLM & -- & -- & -- & Text \\
CoT & -- & -- & -- & Text \\
CoC & -- & \cmark & -- & Program \\
\midrule
RAG$_{\text{NL}}$ & \cmark & -- & -- & Text \\
RAG$_{\text{Code}}$ & \cmark & -- & -- & Text \\
IRCoT & \cmark & -- & \cmark & Text \\
Ours$_{\text{NL}}$ & \cmark & \cmark & \cmark & Program \\
\textbf{Ours} & \cmark & \cmark & \cmark & Program \\
\bottomrule
\end{tabular}
\vspace{-3mm}
\end{table}

\begin{table*}[t]
\centering
\caption{Benchmarks, per-dataset accuracy, and summary statistics of execution outcomes for our iterative codification method.
$N$ denotes the number of program refinement iterations in our loop (average/maximum over instances).}
\vspace{-3mm}

\label{tab:main_combined}
\footnotesize
\setlength{\tabcolsep}{2.2pt}
\renewcommand{\arraystretch}{0.92}
\resizebox{\textwidth}{!}{%
\begin{tabular}{l c | r r| r r r r r |r  ||  c c c}
\toprule
\multicolumn{2}{c|}{\textbf{Benchmarks}} &
\multicolumn{8}{c||}{\textbf{Accuracy}} &
\multicolumn{3}{c}{\textbf{Execution outcomes (Ours)}} \\
\cmidrule(lr){1-2}\cmidrule(lr){3-10}\cmidrule(lr){11-13}
\textbf{Dataset} & \textbf{Know-} &
\textbf{LLM} & \textbf{CoT} &
\multicolumn{5}{c|}{\textbf{Ablations}} &
\textbf{ Ours } &
\textbf{Exec-} & \textbf{Exception} & \textbf{$N$} \\
& \textbf{ledge} &
& &
\textbf{CoC} & \textbf{RAG$_{\text{NL}}$} & \textbf{RAG$_{\text{Code}}$} & \textbf{IRCoT} & \textbf{Ours$_{\text{NL}}$} &\name{}
& \textbf{utability(\%)} & \textbf{Ratio(\%)} & (Avg/Max) \\
\midrule
\multicolumn{13}{c}{Mathematical reasoning and proof-like benchmarks} \\
\midrule
GSM8K        & None  & 0.93 & 0.97 & 0.96 & 0.96 & 0.89 & 0.94 & 0.98 & \textbf{0.98} & 99 & 0.21 & 1.09/2 \\
ProofNet     & None  & 0.79 & 0.88 & 0.90 & 0.76 & 0.84 & 0.93 & 0.94 & \textbf{0.95} & 99 & 0.43 & 1.73/5 \\
\midrule
\multicolumn{13}{c}{Causal reasoning benchmark} \\
\midrule
IFQA         & Wiki  & 0.32 & 0.31 & 0.34 & 0.35 & 0.41 & 0.36 & 0.42 & \textbf{0.51} & 98 & 0.50 & 1.25/3 \\
Corr2Cause   & None  & 0.65 & 0.81 & 0.79 & 0.59 & 0.74 & 0.71 & 0.80 & \textbf{0.86} & 99 & 0.92 & 2.01/7 \\
\midrule
\multicolumn{13}{c}{Big Bench Hard/Extra Hard} \\
\midrule
BBH          & None  & 0.64 & 0.68 & 0.71 & 0.66 & 0.68 & 0.76 & 0.81 & \textbf{0.88} & 88 & 10.36 & 1.14/3 \\
BBEH         & None  & 0.17 & 0.16 & 0.25 & 0.19 & 0.18 & 0.20 & 0.28 & \textbf{0.37} & 86 & 12.72 & 1.24/4 \\
\midrule
\multicolumn{13}{c}{Legal benchmarks} \\
\midrule
CAIL         & Cases & 0.54 & 0.54 & 0.66 & 0.69 & 0.66 & 0.61 & 0.67 & \textbf{0.72} & 94 & 5.15 & 1.77/5 \\
ECHR         & HUDOC & 0.51 & 0.53 & 0.55 & 0.60 & 0.48 & 0.61 & 0.62 & \textbf{0.64} & 87 & 12.20 & 2.38/9 \\
CAP          & USC   & 0.57 & 0.61 & 0.55 & 0.58 & 0.55 & 0.55 & 0.65 & \textbf{0.68} & 85 & 14.11 & 2.12/8 \\
\midrule
\multicolumn{13}{c}{Medical benchmarks} \\
\midrule
PubMedQA     & TB    & 0.67 & 0.69 & 0.57 & 0.84 & 0.70 & 0.78 & 0.84 & \textbf{0.93} & 92 & 5.72 & 1.62/4 \\
USMLE        & TB    & 0.80 & 0.81 & 0.61 & 0.82 & 0.82 & 0.83 & 0.90 & \textbf{0.91} & 98 & 1.77 & 2.03/7 \\
Health-Claim & PM    & 0.57 & 0.65 & 0.61 & 0.64 & 0.59 & 0.65 & 0.67 & \textbf{0.71} & 98 & 0.49 & 1.02/2 \\
BioASQ-6B    & MED   & 0.73 & 0.75 & 0.79 & 0.81 & 0.66 & 0.79 & 0.84 & \textbf{0.92} & 96 & 3.43 & 1.16/3 \\
\bottomrule
\end{tabular}}
\vspace{1mm}

\begin{minipage}{\textwidth}
\footnotesize
KB: None, Wiki (Wikipedia), HUDOC (ECHR documents), USC (U.S.\ Code),
TB (textbooks), PM (PubMed abstracts), MED (MEDLINE).
\end{minipage}
\vspace{-7mm}
\end{table*}

\subsection{Baselines}
\label{sec:baselines}

We compare against baselines that isolate three design factors:
\textbf{retrieval}, \textbf{execution}, and \textbf{iteration}
(Table~\ref{tab:baseline}).
LLM and CoT (rows 1--2) are text-only prompting baselines.
CoC (row 3) generates a single Python program and executes it once, without retrieval or iterative revision.

To test retrieval without execution, RAG$_{\textsc{NL}}$ and RAG$_{\textsc{Code}}$ (rows 4--5) retrieve evidence but output the final answer as text.
IRCoT (row 6) interleaves retrieval with multi-step \emph{textual} reasoning, still without program execution.
Ours$_{\textsc{NL}}$ and Ours (rows 7--8) run the same iterative \emph{executable} refinement loop; they differ only in the form of retrieved $R_2$ exemplars:
Ours$_{\textsc{NL}}$ conditions revision on natural-language exemplars, whereas Ours uses codified snippet exemplars.
\vspace{-0.3em}
\subsection{Prompt Templates and Computational Overhead}
\label{sec:repro}
\vspace{-0.3em}
{We report in Appendix~A the prompt templates used in this study, including the initial computable-generation prompt and the iterative revision prompt.
We also report there the main inference-time procedure, including retrieval configuration, the stopping condition for the refinement loop, and the maximum number of refinement rounds.
Representative computables and execution traces are provided in Appendix~B.}

{Our method requires additional computational cost relative to the simpler baselines.
For \textsc{CoT} and \textsc{CoC}, each instance is handled by a single main LLM call.
In contrast, our method uses one initial codification call followed by iterative revision, so the total number of main model calls grows with the number of refinement rounds.
Accordingly, the statistic \(N\) reported in Table~\ref{tab:main_combined} (last column) serves as a rough estimate of the additional call-level overhead.
In addition, each call is typically longer than in the text-only baselines because it includes executable codification instructions, retrieved contextual information, and revision feedback.
This additional cost is justified by the gains in accuracy and inspectability reported below.}


\vspace{-0.3em}
\section{Task Accuracy}
\vspace{-0.3em}
We first examine benchmark accuracy following each
dataset's standard protocol.
For codifying methods, answers are
obtained by executing the generated program; for text-only methods,
answers are extracted from model outputs according to dataset-specific
formats.

\label{sec:mainresults}

Table~\ref{tab:main_combined} reports per-dataset accuracy for all methods (LLM/CoT, ablations, and \textbf{Ours}, accuracy column in the middle).
Overall, \textbf{Ours} achieves the strongest accuracy across benchmarks.
The gains are clearest in rule- and exception-heavy domains (legal and biomedical), where correct decisions depend on recovering implicit applicability conditions and aligning them with protocol constraints.
We also see large improvements on broad multi-step suites (BBH: 0.64$\rightarrow$0.88; BBEH: 0.17$\rightarrow$0.37).

{The ablations in Table~\ref{tab:main_combined} support three
  main conclusions.  First, transforming the task into code already
  improves accuracy, but this alone is not sufficient.  Across several
  datasets, \textsc{CoC} improves over \textsc{CoT}, for example on
  BBH (0.68$\rightarrow$0.71), BBEH (0.16$\rightarrow$0.25), CAIL
  (0.54$\rightarrow$0.66), and ProofNet (0.88$\rightarrow$0.90).  This
  improvement suggests that executable codification already helps make
  variables, conditions, and step order explicit.  However, one-shot
  executable codification remains fragile when key conditions are
  still missing or underspecified, which is why \textsc{CoC} alone
  does not reach the best performance.}

\blue{ Second, retrieval contributes most on benchmarks whose correct
  answers depend on background rules or domain knowledge.  For example
  in the legal and biomedical datasets, retrieval-based ablations
  improve over the text-only baselines, for example on CAIL, ECHR,
  CAP, PubMedQA, and BioASQ-6B.  At the same time, purely textual
  retrieval variants remain weaker than the strongest executable
  variants.  This indicates that retrieved evidence alone is often not
  sufficient unless it is turned into an explicit, checkable
  procedure.}

\blue{Third, the strong performance of \textsc{Ours}$_{\textsc{NL}}${(our proposed method without code snippets in RAG)} shows that part of the gain comes from iterative, retrieval-guided restructuring, not only from final code execution.
  For instance, \textsc{Ours}$_{\textsc{NL}}$
  already outperforms the non-iterative baselines on most datasets and is often close to the full method.
The full method is strongest overall when these components are combined: retrieval adds missing information, executable codification turns it into explicit checks, and iteration revises the current program when conditions remain missing or underspecified.}

\blue{The same overall tendency is also supported by the cross-model results reported in Appendix~D, although the absolute gains remain model-dependent.}

\section{Analyses of Benchmark Language under Weakened Formal Semantics}
\label{sec:computables}

We analyze induced executable programs as semantic evidence under weakened formal semantics using three complementary views:
(i) execution outcomes (Executability, Exception Ratio, and refinement iterations $N$),
(ii) procedural behavior under controlled input variation, and
(iii) the explicit tests and branching patterns expressed in code (control-flow and AST summaries).

As references, we compare the induced programs with three publicly released Python corpora (MBPP~\citep{MBPP}, Python-Stack~\citep{PythonStack}, Python-Reasoning~\citep{PythonReasoning}), which provide natural baselines for human-authored code.

\subsection{Executability}
\label{sec:executability}

First, we investigate executability under a stronger formalization regime using the Lean theorem prover \citep{demoura2015lean}, one of the most widely used proof assistants. For each task, we manually design a task-specific Lean template that fixes (i) the theorem statement encoding the instance-to-label mapping and (ii) a proof skeleton with explicit placeholders to be instantiated per instance. The LLM is then asked to fill these placeholders and produce a complete \texttt{.lean} script, which is subsequently checked by Lean.

\blue{This comparison is not intended as a full formal-semantics baseline.
It is a feasibility probe of whether benchmark instances can be instantiated in a stronger proof-oriented environment.
In this setup, the template fixes the overall theorem-and-proof structure, and the LLM only fills the instance-specific content.}

\blue{Failures arise from two main sources.
First, some instances require assumptions or intermediate lemmas that are difficult to encode in the fixed template.
Second, some benchmark descriptions are too open-textured to map cleanly onto the formal skeleton.
 Detailed Lean construction and examples are provided in Appendix E}

We record the fraction of sampled instances whose generated script is accepted by Lean without errors. This fraction is highly task-dependent (e.g., BBH-Dyck: 100\%, GSM8K: 69\%, CAIL: 11\%), indicating a sharp feasibility gap as domain assumptions and open-textured language increase. For some benchmarks, the success rate is too low to treat proof-level formalization as a feasible baseline at scale.

In contrast, Table~\ref{tab:main_combined} (right-most columns) reports three summary statistics of execution outcomes for \name{}.
\emph{Executability} (3\textsuperscript{rd} column to the right) is the fraction of instances whose final program $c^\star$ completes without a runtime error.
\emph{Exception Ratio} (2\textsuperscript{nd} column to the right) is the fraction of instances whose final program raises a runtime error; it captures a stricter failure mode that remains even after revision.
Finally, $N$ (last column; Avg/Max) is the number of compile--execute--feedback cycles used by the loop.

Overall, executability is generally high with a small exception ratio, and $N$ often exceeds 1 on nontrivial benchmarks, suggesting that the approach produces runnable programs and that iteration performs substantive refinement rather than a single-pass translation.

Across benchmark families, harder domains tend to show lower Executability, higher Exception Ratio, and larger $N$.
Importantly, $N$ increasing with difficulty is also evidence that the iteration loop is doing nontrivial work: when implicit conditions are dense, the system repeatedly uses execution feedback to identify missing checks and revises the program until it becomes runnable.
The remaining Exception Ratio highlights a residual set of instances where benchmark conditions are still not stably compilable into executable checks under our protocol, marking practical limits of codification in these settings.

\begin{table}[t]
\centering
\footnotesize
\caption{Metamorphic-test accuracy (50 programs per dataset).
\textbf{Metamorphic-test accuracy} is the accuracy of re-running the \emph{same} program on newly constructed validity-preserving inputs.}
\label{tab:reinput}
\vspace{-0.2cm}
\setlength{\tabcolsep}{4pt}
\begin{tabular}{lcc}
\toprule
\textbf{Dataset} & \textbf{Accuracy} & \textbf{\shortstack{Metamorphic-test\\ accuracy}} \\
\midrule
GSM8K & 0.98 & 1.00 \\
Corr2Cause & 0.86 & 0.97 \\
BBH Arithmetic & 1.00 & 1.00 \\
BBH Dyck & 0.83 & 1.00 \\
BBEH Bool Expr & 0.99 & 1.00 \\
\midrule
\textit{Avg.} & \textit{0.91} & \textit{0.96} \\
\bottomrule
\end{tabular}
\vspace{-2mm}
\end{table}
\subsection{Procedural Behavior under Input Variation}
\label{sec:reinput}
Instance-level accuracy tests whether the system outputs the correct label for the given instance, but it does not test whether the induced program behaves as a reusable procedure.
We therefore evaluate procedural behavior using \textbf{metamorphic testing} \citep{meta}:
we construct validity-preserving follow-up inputs manually and re-run the \emph{same} compiled program to check whether it continues to produce correct outputs.

{Metamorphic-test accuracy} is computed on a subset of executable programs.
For each dataset in Table~\ref{tab:reinput}, we sample 50 instances whose final programs execute successfully, manually construct two validity-preserving follow-up inputs per program, and re-run the fixed program on these inputs.
\blue{The resulting outputs are then checked manually against the expected outputs for the modified inputs.}
This metric is not meant to be directly comparable to instance-level Accuracy (it is conditioned on executability and evaluated on a sample);
instead, it probes whether a compiled program generalizes as a procedure beyond its original instance.

We apply this probe only to benchmarks where valid input variation can be constructed unambiguously (e.g., arithmetic/algorithmic tasks, Dyck/Boolean-expression tasks, and causal tasks with controlled relation changes).
High metamorphic-test accuracy on these datasets indicates that many induced programs behave like reusable procedures rather than one-off scripts, strengthening the case for treating executable programs as semantic evidence rather than merely answer-producing tools.

\begin{figure}[t]
\centering
\includegraphics[width=\linewidth]{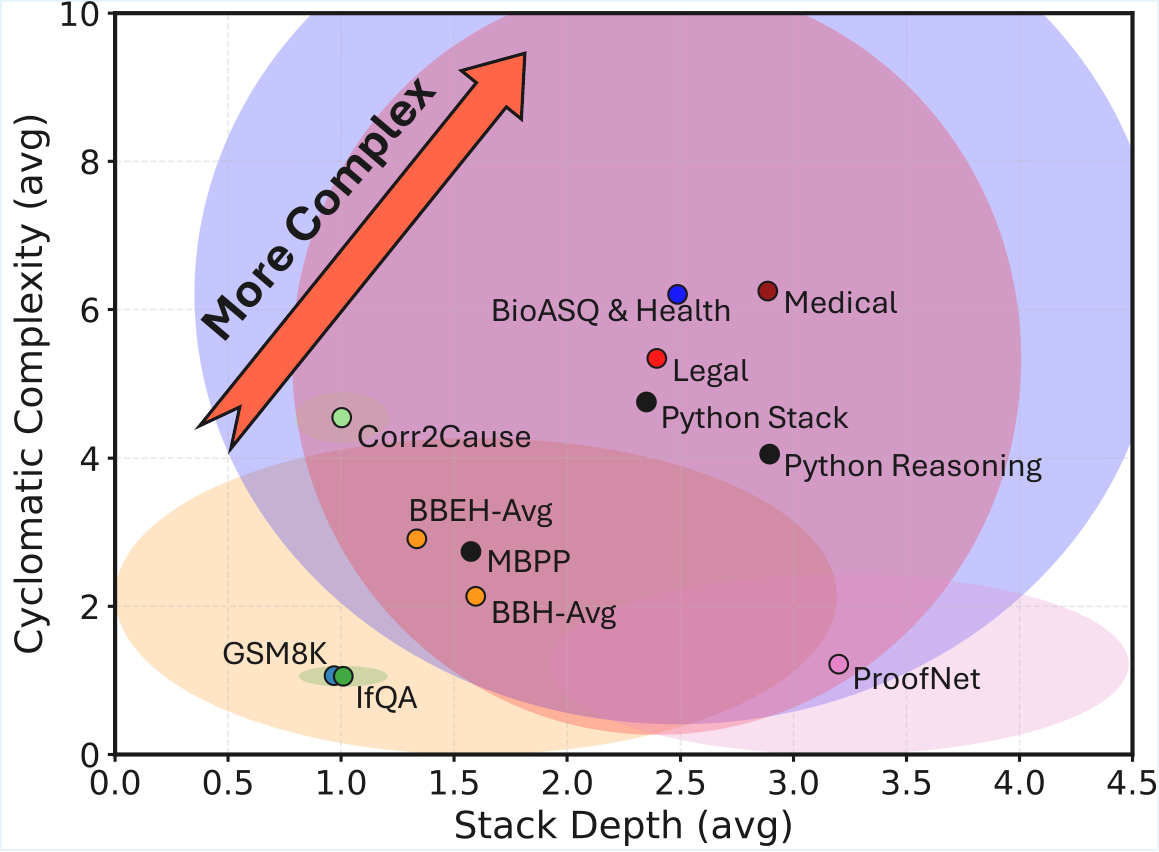}
\vspace{-6mm}
\caption{Stack depth vs.\ cyclomatic complexity across task families. Points include benchmark-induced computables and reference Python corpora (used only for structural calibration).}
\label{fig:cf_complexity}
\vspace{-5mm}
\end{figure}


\begin{figure*}[t]
\centering
\begin{minipage}[b]{0.5\textwidth}
  \centering
  \includegraphics[width=\linewidth]{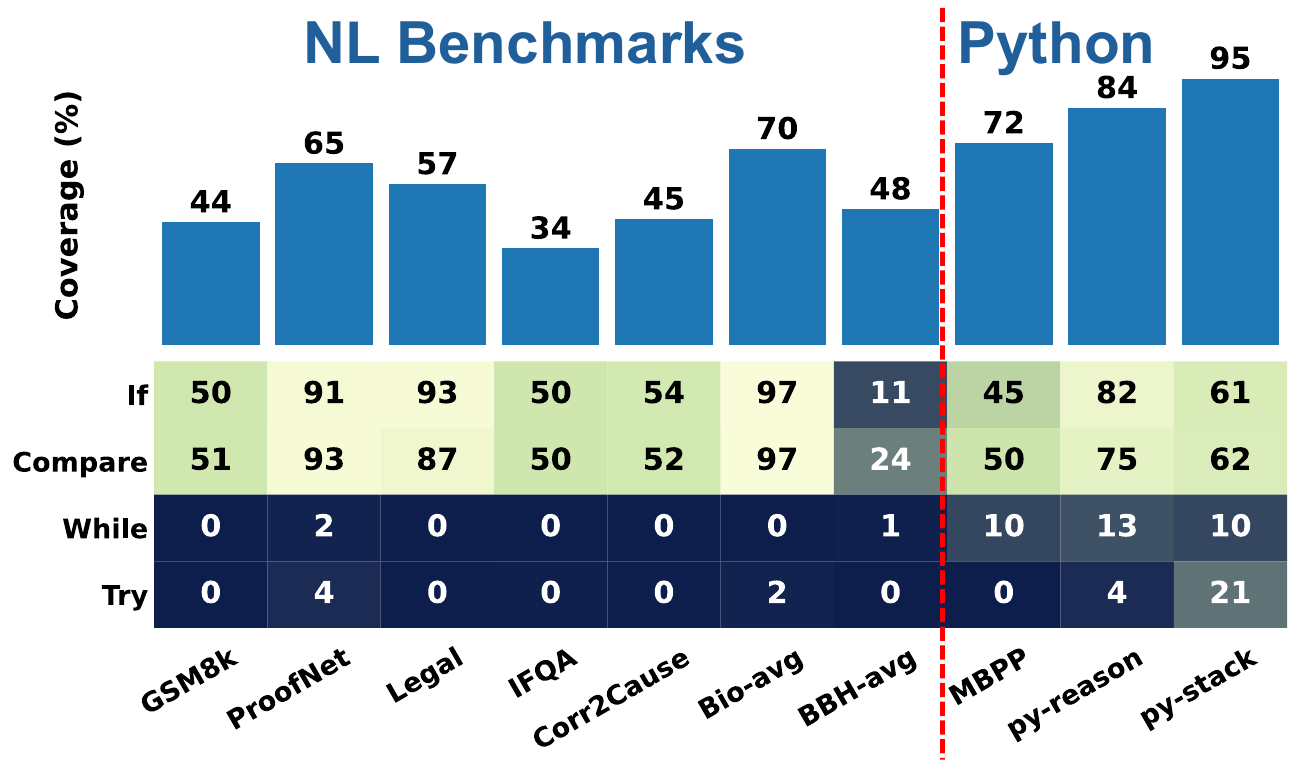}\\[-1mm]
  \footnotesize\textbf{(a)} AST footprint (coverage + presence).
\end{minipage}
\begin{minipage}[b]{0.49\textwidth}
  \centering
  \includegraphics[width=\linewidth]{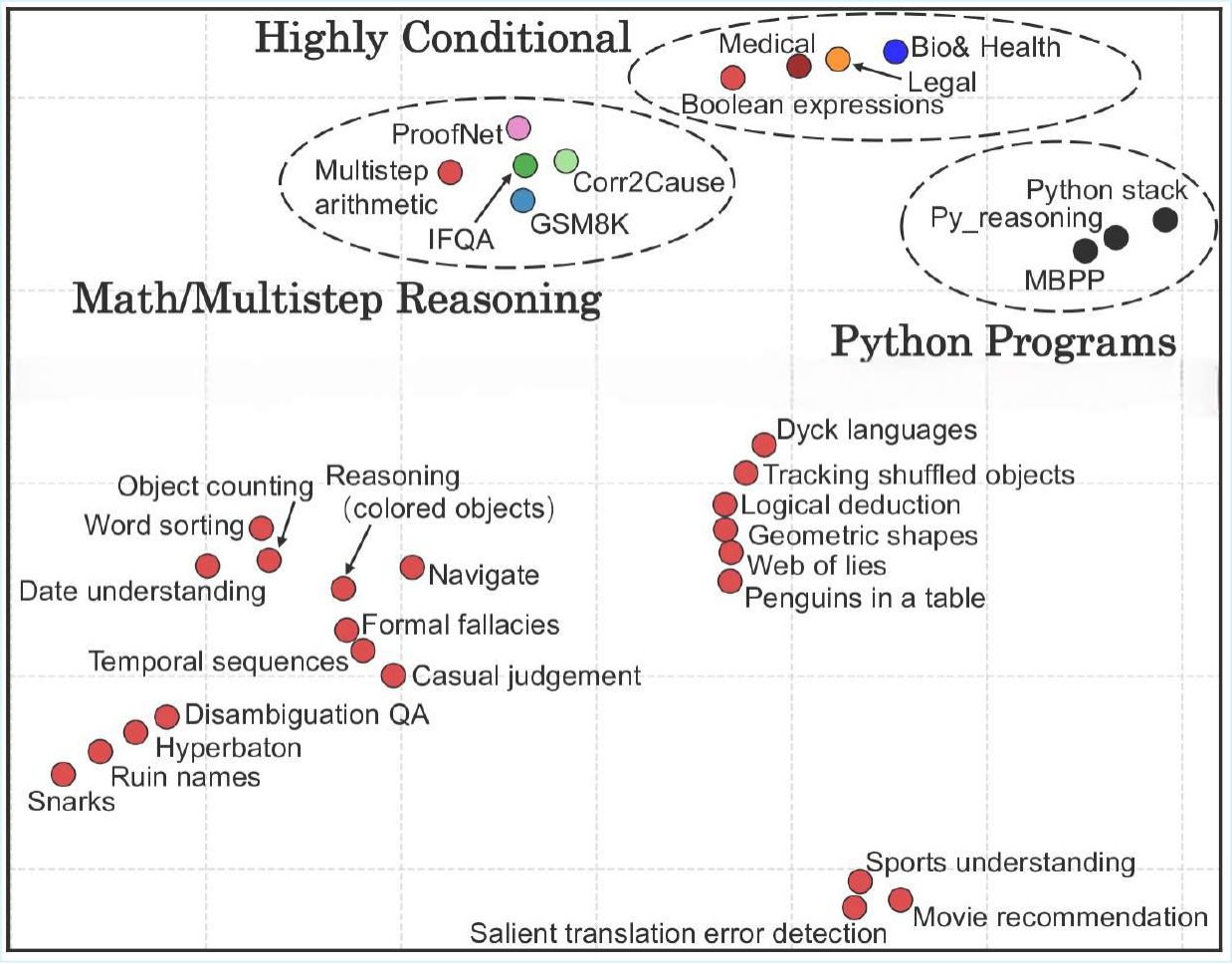}\\[-1mm]
  \footnotesize\textbf{(b)} t-SNE of AST-profile embeddings.
\end{minipage}\hfill

\vspace{-1mm}
\caption{AST-structure views of benchmark-induced computables.
(a) AST footprint: node-type coverage (top) and presence rates of \texttt{If}, \texttt{Compare}, \texttt{While}, \texttt{Try} (bottom).
(b) t-SNE of AST-profile embeddings; red points denote BBEH categories/subtasks.}
\label{fig:ast_views}
\vspace{-5mm}
\end{figure*}

\subsection{Complexity of Control-Flow Structure}
\label{sec:complexity}

To characterize the control-flow structure of induced computables, we analyze stack depth and cyclomatic complexity~\cite{1702388} of the induced programs, capturing nesting and branching, respectively.
Figure~\ref{fig:cf_complexity} plots these measures across benchmark families.

The resulting programs exhibit systematic differences across domains.
GSM8K and IFQA concentrate in the low-depth/low-complexity region, consistent with largely straight-line computation with few conditional branches.
Corr2Cause shows relatively high cyclomatic complexity at modest stack depth, suggesting frequent case splits (many \texttt{if}/\texttt{elif} branches) without deep nesting.
ProofNet exhibits the opposite pattern: high depth but low cyclomatic complexity, consistent with deep, stepwise verification or construction rather than extensive branching.

Legal and Bio/Medical occupy the upper-right region (high depth and high branching), indicating that they require both (i) composing multiple conditions and (ii) separating applicability boundaries and exceptions via explicit branches.

This structural pattern is consistent with their larger refinement counts $N$ in Table~\ref{tab:main_combined}:
domains that require more revisions also tend to induce programs with richer branching/nesting, because iteration turns missing conditions revealed by execution into additional explicit checks in code.

\subsection{AST-based Patterns}
\label{sec:ast_patterns}

We analyze the induced programs via their Python abstract syntax trees (ASTs).
For each program, we form an \emph{AST profile} that records which node types appear.
This yields a lightweight, inspectable summary of what executable constructs benchmark language repeatedly induces.

Figure~\ref{fig:ast_views}(a) compares AST profiles across benchmark families and reference Python corpora.
First, benchmark-induced programs draw from a less AST vocabulary than the reference corpora
(coverage $34$--$70\%$ vs.\ $72$--$95\%$),
suggesting that current benchmarks repeatedly elicit a constrained subset of executable structures.
Second, Legal and Bio show consistently high \texttt{If} and \texttt{Compare} presence rates,
indicating that their induced programs frequently check preconditions (via comparisons) and branch to handle different cases.

Figure~\ref{fig:ast_views}(b) visualizes AST-profile similarity with t-SNE, which we read as a qualitative structural map.
The embedding suggests two dominant regimes.
Points near the \emph{Highly Conditional} region include Legal and Bio/Health, consistent with programs dominated by many condition checks that must all be satisfied (dense Boolean gating in \texttt{If}/\texttt{Compare} and related nodes; Fig.~\ref{fig:ast_views}(a)).
In contrast, the \emph{Math/Multistep Reasoning} region contains GSM8K, ProofNet, and several BBH subtasks, reflecting a more stepwise procedural style that is less dominated by dense guard structure.
Notably, IFQA and Corr2Cause also lie in this region, suggesting that some causal instances are solved with program structures closer to multi-step construction/checking than to heavy case-splitting.

From another perspective, these structural summaries allow benchmark families to be compared at the level of procedural commitments,
i.e., which executable checks and control-flow constructs the benchmark language repeatedly forces the model to commit to.
In Figure~\ref{fig:ast_views}(b), red points denote individual BBEH categories, while the other colored points denote broader benchmark or domain families.
The left set of red BBEH points corresponds mainly to natural-language reasoning categories such as date understanding, word sorting, object counting, formal fallacies, and causal judgement.
The middle-right set corresponds to more structured state-tracking or formal-control categories such as Dyck languages, tracking shuffled objects, logical deduction, geometric shapes, web of lies, and penguins in a table.
The low-density regions between these sets suggest that existing suites more often elicit either dense Boolean gating with many guards or stepwise construction with relatively light case-splitting, but less often programs that combine both.
This points to a benchmark-design direction: construct tasks that force these missing mixtures of control-flow and test constructs, rather than repeatedly eliciting the same templates.

\blue{This interpretation is also consistent with the cross-model results in Appendix~D, which show that the main pattern transfers beyond the primary GPT-5-based setting while remaining sensitive to model quality.}

\section{Implications}

Our findings have two implications: for benchmark design and evaluation, and for the empirical viability of weakened formal semantics as a way to interpret benchmark language through executable representations.

\paragraph{Implications for benchmark design.}
In Section~\ref{sec:computables}, our structural diagnostics show that
various benchmark families differ in the kinds of procedures they
elicit. Some families repeatedly elicit dense multi-condition checks
(explicit applicability tests), while others rely on shallower, more
linear routines. Reporting these execution and structural signals
alongside accuracy can help with showing what kinds of procedures a
benchmark tends to require, and clarify why similar accuracy numbers
may correspond to meaningfully different executable
representations. \blue{This observation is also relevant to benchmark validity: beyond aggregate accuracy, execution outcomes and structural properties of computables help characterize what kinds of procedures benchmark instances are actually eliciting \citep{bean2025construct}.}

The t-SNE map also shows clear gaps in the observed AST-profile space, suggesting that current benchmark families elicit an uneven subset of executable structures; these gaps point to opportunities for future benchmarks that target under-represented procedural patterns.
\paragraph{Implications for weakened formal semantics.}
Beyond evaluation, our results support weakened formal semantics as a practical way to study interpretation when proof-level guarantees are out of reach.
Instead of treating intermediate reasoning as pure text, weakened formal semantics treats an induced executable representation as a minimal semantic witness: it can be run, probed under controlled input changes, and revised when execution reveals errors or missing conditions.
Although these witnesses are not solid proofs, they make the basis of an answer explicit in a form that supports targeted debugging and comparison across domains.
Our experiments suggest that this form of accountability is viable at benchmark scale when retrieval supplies missing contents and iteration turns feedback into targeted repairs.

\section{Conclusion}
\label{sec:conclusion}

We introduced \emph{weakened formal semantics} as a practical stance for interpreting benchmark language without needing complete logical formalization or proof guarantees.
Our core claim is that, in task-oriented settings, semantic adequacy can be studied through induced \emph{computables} whose runtime behavior provides operational evidence of what an interpretation commits to.

We instantiated this stance with a retrieval-grounded, iterative codification loop that compiles task descriptions into Python computables and refines them using retrieved evidence and feedback.
Across diverse benchmarks, this approach improves over text-only prompting and one-shot code execution, while producing runnable representations that can be inspected and stress-tested rather than merely read.
In doing so, it not only increases accuracy, but also turns implicit requirements (conditions, procedures, exceptions) into explicit, checkable steps.

Finally, we used summary statistics of execution outcomes, control-flow complexity, and AST-based summaries to compare benchmarks by the kinds of procedures they repeatedly elicit.
Our goal is not to replace proof-checked formalization, but to provide an empirical bridge: a scalable way to reveal where benchmark language \emph{can} be compiled into explicit procedures and where it \emph{resists} such compilation.
This makes practical limits of formalization observable in terms of runnable behavior, and offers a concrete basis for diagnosing benchmarks beyond final answers.

\section{Limitations}

Our study has several limitations and points to directions for future work.
First, the approach induces Python programs, which biases the analysis toward tasks whose procedural content can be expressed in an imperative language.
Although Python is a practical target for LLM-based procedural translation---given its ubiquity, permissive syntax, and tool ecosystem---this choice also narrows the representational lens.
Some meanings, including graded judgments, underspecified discourse relations, defeasible interpretations, and heavily context-dependent inferences, are not naturally captured in this form.
More broadly, computables privilege interpretations that can be externalized as explicit procedures.

\blue{Second, executability should not be equated with semantic correctness: a computable may execute successfully while still encoding an incomplete, overly instance-specific, or option-sensitive procedure.}

\blue{At the same time, executability can still serve as a diagnostic signal.
Because the intermediate procedure is explicit and inspectable, it can help reveal omitted conditions, branches that are too closely tied to answer options, and checks that fail to capture the intended applicability constraints.}

Third, retrieval and iteration add computational cost and depend on the quality of external sources.
Although we control for exemplar leakage, performance and stability may vary with different corpora, retrieval strategies, or revision policies.
We report additional implementation details, model settings, and supplementary refinement-related statistics in the appendix, but further work is needed to characterize the cost--benefit trade-off of iterative executable refinement more systematically.

Finally, our analysis targets benchmark-style task descriptions.
Whether similar regularities hold for open-ended discourse, interactive dialogue, or longer-horizon language use remains unclear.
Extending weakened formal semantics beyond benchmark settings is therefore an important direction for future work.

\section*{Acknowledgments}

This work was supported by JST CREST, Japan, Grant Number JPMJCR2114, and JST BOOST, Japan, Grant Number JPMJBS2429.

\sloppy
\bibliography{tacl}

\begin{thebibliography}{52}
\expandafter\ifx\csname natexlab\endcsname\relax\def\natexlab#1{#1}\fi

\bibitem[{Azerbayev et~al.(2023)Azerbayev, Piotrowski, Schoelkopf, Ayers, Radev, and Avigad}]{proofnet}
Zhangir Azerbayev, Bartosz Piotrowski, Hailey Schoelkopf, Edward~W. Ayers, Dragomir Radev, and Jeremy Avigad. 2023.
\newblock \href {http://arxiv.org/abs/2302.12433} {Proofnet: Autoformalizing and formally proving undergraduate-level mathematics}.

\bibitem[{Barwise and Cooper(1981)}]{Barwise1981Generalized}
Jon Barwise and Robin Cooper. 1981.
\newblock \href {https://doi.org/10.1007/bf00350139} {Generalized quantifiers and natural language}.
\newblock \emph{Linguistics and Philosophy}, 4(2):159--219.

\bibitem[{Bean et~al.(2025)}]{bean2025construct}
Andrew~M. Bean et~al. 2025.
\newblock \href {https://openreview.net/forum?id=mdA5lVvNcU} {Measuring what matters: Construct validity in large language model benchmarks}.
\newblock In \emph{The Thirty-Ninth Annual Conference on Neural Information Processing Systems Datasets and Benchmarks Track}.

\bibitem[{Bertot and Cast{\'e}ran(2004)}]{bertot2004coqart}
Yves Bertot and Pierre Cast{\'e}ran. 2004.
\newblock \emph{Interactive Theorem Proving and Program Development: {Coq'Art}: The Calculus of Inductive Constructions}.
\newblock Springer.

\bibitem[{Besta et~al.(2024)Besta, Blach, Kubicek, Gerstenberger, Podstawski, Gianinazzi, Gajda, Lehmann, Niewiadomski, Nyczyk, and Hoefler}]{10.1609/aaai.v38i16.29720}
Maciej Besta, Nils Blach, Ales Kubicek, Robert Gerstenberger, Micha\l{} Podstawski, Lukas Gianinazzi, Joanna Gajda, Tomasz Lehmann, Hubert Niewiadomski, Piotr Nyczyk, and Torsten Hoefler. 2024.
\newblock \href {https://doi.org/10.1609/aaai.v38i16.29720} {Graph of thoughts: solving elaborate problems with large language models}.
\newblock In \emph{Proceedings of the Thirty-Eighth AAAI Conference on Artificial Intelligence and Thirty-Sixth Conference on Innovative Applications of Artificial Intelligence and Fourteenth Symposium on Educational Advances in Artificial Intelligence}, AAAI'24/IAAI'24/EAAI'24. AAAI Press.

\bibitem[{Blackburn and Bos(2005)}]{blackburn2005representation}
Patrick Blackburn and Johan Bos. 2005.
\newblock \emph{Representation and Inference for Natural Language: A First Course in Computational Semantics}.
\newblock CSLI Publications.

\bibitem[{Borgeaud et~al.(2022)Borgeaud, Mensch, Hoffmann, Cai, Rutherford, Millican, Van Den~Driessche, Lespiau, Damoc, Clark, De~Las~Casas, Guy, Menick, Ring, Hennigan, Huang, Maggiore, Jones, Cassirer, Brock, Paganini, Irving, Vinyals, Osindero, Simonyan, Rae, Elsen, and Sifre}]{borgeaud2022improvinglanguagemodelsretrieving}
Sebastian Borgeaud, Arthur Mensch, Jordan Hoffmann, Trevor Cai, Eliza Rutherford, Katie Millican, George~Bm Van Den~Driessche, Jean-Baptiste Lespiau, Bogdan Damoc, Aidan Clark, Diego De~Las~Casas, Aurelia Guy, Jacob Menick, Roman Ring, Tom Hennigan, Saffron Huang, Loren Maggiore, Chris Jones, Albin Cassirer, Andy Brock, Michela Paganini, Geoffrey Irving, Oriol Vinyals, Simon Osindero, Karen Simonyan, Jack Rae, Erich Elsen, and Laurent Sifre. 2022.
\newblock \href {https://proceedings.mlr.press/v162/borgeaud22a.html} {Improving language models by retrieving from trillions of tokens}.
\newblock In \emph{Proceedings of the 39th International Conference on Machine Learning}, volume 162 of \emph{Proceedings of Machine Learning Research}, pages 2206--2240. PMLR.

\bibitem[{Bos(2008)}]{bos2008wide}
Johan Bos. 2008.
\newblock Wide-coverage semantic analysis with boxer.
\newblock In \emph{Semantics in Text Processing. STEP 2008 Conference Proceedings}, pages 277--286. College Publications.

\bibitem[{Chalkidis et~al.(2021)Chalkidis, Fergadiotis, Tsarapatsanis, Aletras, Androutsopoulos, and Malakasiotis}]{ECHR}
Ilias Chalkidis, Manos Fergadiotis, Dimitrios Tsarapatsanis, Nikolaos Aletras, Ion Androutsopoulos, and Prodromos Malakasiotis. 2021.
\newblock \href {https://doi.org/10.18653/v1/2021.naacl-main.22} {Paragraph-level rationale extraction through regularization: A case study on {E}uropean court of human rights cases}.
\newblock In \emph{Proceedings of the 2021 Conference of the North American Chapter of the Association for Computational Linguistics: Human Language Technologies}, pages 226--241, Online. Association for Computational Linguistics.

\bibitem[{Chen et~al.(2018)Chen, Kuo, Liu, Poon, Towey, Tse, and Zhou}]{meta}
Tsong~Yueh Chen, Fei-Ching Kuo, Huai Liu, Pak-Lok Poon, Dave Towey, T.~H. Tse, and Zhi~Quan Zhou. 2018.
\newblock \href {https://doi.org/10.1145/3143561} {Metamorphic testing: A review of challenges and opportunities}.
\newblock \emph{ACM Comput. Surv.}, 51(1).

\bibitem[{Chen et~al.(2023)Chen, Ma, Wang, and Cohen}]{chen2023programthoughtspromptingdisentangling}
Wenhu Chen, Xueguang Ma, Xinyi Wang, and William~W. Cohen. 2023.
\newblock \href {https://openreview.net/forum?id=YfZ4ZPt8zd} {Program of thoughts prompting: Disentangling computation from reasoning for numerical reasoning tasks}.
\newblock \emph{Transactions on Machine Learning Research}.

\bibitem[{Cobbe et~al.(2021)Cobbe, Kosaraju, Bavarian, Chen, Jun, Kaiser, Plappert, Tworek, Hilton, Nakano, Hesse, and Schulman}]{gsm8k}
Karl Cobbe, Vineet Kosaraju, Mohammad Bavarian, Mark Chen, Heewoo Jun, Lukasz Kaiser, Matthias Plappert, Jerry Tworek, Jacob Hilton, Reiichiro Nakano, Christopher Hesse, and John Schulman. 2021.
\newblock \href {http://arxiv.org/abs/2110.14168} {Training verifiers to solve math word problems}.

\bibitem[{Frege(1879)}]{frege}
Gottlob Frege. 1879.
\newblock Begriffsschrift, a formula language, modeled upon that of arithmetic, for pure thought [1879].
\newblock \emph{From Frege to G\"{o}del: A Source Book in Mathematical Logic}, 1931:1--82.

\bibitem[{Gao et~al.(2023)Gao, Madaan, Zhou, Alon, Liu, Yang, Callan, and Neubig}]{10.5555/3618408.3618843}
Luyu Gao, Aman Madaan, Shuyan Zhou, Uri Alon, Pengfei Liu, Yiming Yang, Jamie Callan, and Graham Neubig. 2023.
\newblock {PAL}: Program-aided {Language Models}.
\newblock In \emph{Proceedings of the 40th International Conference on Machine Learning}, ICML'23. JMLR.org.

\bibitem[{{Google Research}(2021)}]{MBPP}
{Google Research}. 2021.
\newblock \href {https://github.com/google-research/google-research/tree/master/mbpp} {{MBPP}: Mostly basic python problems}.
\newblock [Accessed 2025-05-12].

\bibitem[{Heim and Kratzer(1998)}]{heim1998semantics}
Irene Heim and Angelika Kratzer. 1998.
\newblock \emph{Semantics in Generative Grammar}.
\newblock Blackwell.

\bibitem[{{Hugging Face Datasets}(2023)}]{PythonStack}
{Hugging Face Datasets}. 2023.
\newblock \href {https://huggingface.co/datasets/PatrickHaller/the-stack-python-1M} {the-stack-python-1m (patrickhaller)}.
\newblock [Accessed 2025-05-12].

\bibitem[{{Hugging Face Datasets}(2024)}]{PythonReasoning}
{Hugging Face Datasets}. 2024.
\newblock \href {https://huggingface.co/datasets/notbadai/python_functions_reasoning} {python\_functions\_reasoning (notbadai)}.
\newblock [Accessed 2025-05-12].

\bibitem[{Izacard et~al.(2023)Izacard, Lewis, Lomeli, Hosseini, Petroni, Schick, Dwivedi-Yu, Joulin, Riedel, and Grave}]{izacard2022atlasfewshotlearningretrieval}
Gautier Izacard, Patrick Lewis, Maria Lomeli, Lucas Hosseini, Fabio Petroni, Timo Schick, Jane Dwivedi-Yu, Armand Joulin, Sebastian Riedel, and Edouard Grave. 2023.
\newblock Atlas: few-shot learning with retrieval augmented language models.
\newblock \emph{J. Mach. Learn. Res.}, 24(1).

\bibitem[{Jiang et~al.(2024)Jiang, Dong, Wang, Fang, Shang, Li, Jin, and Jiao}]{jiang}
Xue Jiang, Yihong Dong, Lecheng Wang, Zheng Fang, Qiwei Shang, Ge~Li, Zhi Jin, and Wenpin Jiao. 2024.
\newblock \href {https://doi.org/10.1145/3672456} {Self-planning code generation with {Large Language Models}}.
\newblock \emph{ACM Trans. Softw. Eng. Methodol.}, 33(7).

\bibitem[{Jimenez et~al.(2024)Jimenez, Yang, Wettig, Yao, Pei, Press, and Narasimhan}]{jimenez2024swebenchlanguagemodelsresolve}
Carlos~E Jimenez, John Yang, Alexander Wettig, Shunyu Yao, Kexin Pei, Ofir Press, and Karthik~R Narasimhan. 2024.
\newblock \href {https://openreview.net/forum?id=VTF8yNQM66} {{SWE}-bench: Can language models resolve real-world github issues?}
\newblock In \emph{The Twelfth International Conference on Learning Representations}.

\bibitem[{Jin et~al.(2019)Jin, Dhingra, Liu, Cohen, and Lu}]{jin2019pubmedqa}
Qiao Jin, Bhuwan Dhingra, Zhengping Liu, William Cohen, and Xinghua Lu. 2019.
\newblock {PubMedQA}: A dataset for biomedical research question answering.
\newblock In \emph{Proceedings of the 2019 Conference on Empirical Methods in Natural Language Processing and the 9th International Joint Conference on Natural Language Processing (EMNLP-IJCNLP)}, pages 2567--2577.

\bibitem[{Jin et~al.(2024)Jin, Liu, Lyu, Poff, Sachan, Mihalcea, Diab, and Sch{\"o}lkopf}]{jin2024largelanguagemodelsinfer}
Zhijing Jin, Jiarui Liu, Zhiheng Lyu, Spencer Poff, Mrinmaya Sachan, Rada Mihalcea, Mona~T. Diab, and Bernhard Sch{\"o}lkopf. 2024.
\newblock \href {https://openreview.net/forum?id=vqIH0ObdqL} {Can large language models infer causation from correlation?}
\newblock In \emph{The Twelfth International Conference on Learning Representations}.

\bibitem[{Johnson et~al.(2021)Johnson, Douze, and Jégou}]{8733051}
Jeff Johnson, Matthijs Douze, and Hervé Jégou. 2021.
\newblock \href {https://doi.org/10.1109/TBDATA.2019.2921572} {{ Billion-Scale Similarity Search with GPUs }}.
\newblock \emph{IEEE Transactions on Big Data}, 7(03):535--547.

\bibitem[{Kamp and Reyle(1993)}]{kamp1993discourse}
Hans Kamp and Uwe Reyle. 1993.
\newblock \emph{From Discourse to Logic: Introduction to Modeltheoretic Semantics of Natural Language, Formal Logic and Discourse Representation Theory}.
\newblock Kluwer Academic Publishers.

\bibitem[{Kazemi et~al.(2025)Kazemi, Fatemi, Bansal, Palowitch, Anastasiou, Mehta, Jain, Aglietti, Jindal, Chen, Dikkala, Tyen, Liu, Shalit, Chiappa, Olszewska, Tay, Tran, Le, and Firat}]{kazemi2025bigbenchextrahard}
Mehran Kazemi, Bahare Fatemi, Hritik Bansal, John Palowitch, Chrysovalantis Anastasiou, Sanket~Vaibhav Mehta, Lalit~K Jain, Virginia Aglietti, Disha Jindal, Peter Chen, Nishanth Dikkala, Gladys Tyen, Xin Liu, Uri Shalit, Silvia Chiappa, Kate Olszewska, Yi~Tay, Vinh~Q. Tran, Quoc~V Le, and Orhan Firat. 2025.
\newblock \href {https://doi.org/10.18653/v1/2025.acl-long.1285} {{{BIG}-Bench Extra Hard}}.
\newblock In \emph{Proceedings of the 63rd Annual Meeting of the Association for Computational Linguistics (Volume 1: Long Papers)}, pages 26473--26501, Vienna, Austria. Association for Computational Linguistics.

\bibitem[{Kotonya and Toni(2020)}]{health}
Neema Kotonya and Francesca Toni. 2020.
\newblock \href {https://www.aclweb.org/anthology/2020.emnlp-main.623} {Explainable automated fact-checking for public health claims}.
\newblock In \emph{Proceedings of the 2020 Conference on Empirical Methods in Natural Language Processing (EMNLP)}, pages 7740--7754, Online. Association for Computational Linguistics.

\bibitem[{Le and Andrzejak(2024)}]{10.1007/s10515-024-00451-y}
Kim~Tuyen Le and Artur Andrzejak. 2024.
\newblock \href {https://doi.org/10.1007/s10515-024-00451-y} {Rethinking {AI} code generation: a one-shot correction approach based on user feedback}.
\newblock \emph{Automated Software Engg.}, 31(2).

\bibitem[{Lewis et~al.(2020)Lewis, Perez, Piktus, Petroni, Karpukhin, Goyal, K\"{u}ttler, Lewis, Yih, Rockt\"{a}schel, Riedel, and Kiela}]{RAG}
Patrick Lewis, Ethan Perez, Aleksandra Piktus, Fabio Petroni, Vladimir Karpukhin, Naman Goyal, Heinrich K\"{u}ttler, Mike Lewis, Wen-tau Yih, Tim Rockt\"{a}schel, Sebastian Riedel, and Douwe Kiela. 2020.
\newblock Retrieval-augmented generation for knowledge-intensive {NLP} tasks.
\newblock In \emph{Advances in Neural Information Processing Systems}, volume~33, pages 9459--9474. Curran Associates, Inc.

\bibitem[{Li et~al.(2024)Li, Liang, Zeng, Chen, Hausman, Sadigh, Levine, Fei-Fei, Xia, and Ichter}]{coc}
Chengshu Li, Jacky Liang, Andy Zeng, Xinyun Chen, Karol Hausman, Dorsa Sadigh, Sergey Levine, Li~Fei-Fei, Fei Xia, and Brian Ichter. 2024.
\newblock {Chain of Code}: Reasoning with a language model-augmented code emulator.
\newblock In \emph{Proceedings of the 41st International Conference on Machine Learning}, volume 235 of \emph{Proceedings of Machine Learning Research}, pages 28259--28277. PMLR.

\bibitem[{McCabe(1976)}]{1702388}
Thomas~J. McCabe. 1976.
\newblock \href {https://doi.org/10.1109/TSE.1976.233837} {A complexity measure}.
\newblock \emph{IEEE Transactions on Software Engineering}, SE-2(4):308--320.

\bibitem[{Montague(1974{\natexlab{a}})}]{montague1973ptq}
Richard Montague. 1974{\natexlab{a}}.
\newblock The proper treatment of quantification in ordinary english.
\newblock In Richmond~H. Thomason, editor, \emph{Formal Philosophy: Selected Papers of Richard Montague}, pages 247--270. Yale University Press, New Haven.
\newblock Originally circulated 1973.

\bibitem[{Montague(1974{\natexlab{b}})}]{montague1970ug}
Richard Montague. 1974{\natexlab{b}}.
\newblock Universal grammar.
\newblock In Richmond~H. Thomason, editor, \emph{Formal Philosophy: Selected Papers of Richard Montague}, pages 222--246. Yale University Press, New Haven.
\newblock Originally circulated 1970.

\bibitem[{de~Moura et~al.(2015)de~Moura, Kong, Avigad, van Doorn, and von Raumer}]{demoura2015lean}
Leonardo de~Moura, Soonho Kong, Jeremy Avigad, Floris van Doorn, and Jakob von Raumer. 2015.
\newblock The {Lean} theorem prover (system description).
\newblock In \emph{International Conference on Automated Deduction (CADE-25)}, Lecture Notes in Computer Science. Springer.

\bibitem[{Nentidis et~al.(2018)Nentidis, Krithara, Bougiatiotis, Paliouras, and Kakadiaris}]{bio}
Anastasios Nentidis, Anastasia Krithara, Konstantinos Bougiatiotis, Georgios Paliouras, and Ioannis Kakadiaris. 2018.
\newblock \href {https://doi.org/10.18653/v1/W18-5301} {Results of the sixth edition of the {B}io{ASQ} challenge}.
\newblock In \emph{Proceedings of the 6th BioASQ Workshop: A Challenge on Large-Scale Biomedical Semantic Indexing and Question Answering}, pages 1--10, Brussels, Belgium. Association for Computational Linguistics.

\bibitem[{Parvez et~al.(2021)Parvez, Ahmad, Chakraborty, Ray, and Chang}]{parvez2021retrievalaugmentedcodegeneration}
Md~Rizwan Parvez, Wasi Ahmad, Saikat Chakraborty, Baishakhi Ray, and Kai-Wei Chang. 2021.
\newblock \href {https://doi.org/10.18653/v1/2021.findings-emnlp.232} {Retrieval augmented code generation and summarization}.
\newblock In \emph{Findings of the Association for Computational Linguistics: EMNLP 2021}, pages 2719--2734, Punta Cana, Dominican Republic. Association for Computational Linguistics.

\bibitem[{Rabinovich et~al.(2017)Rabinovich, Stern, and Klein}]{rabinovich-etal-2017-abstract}
Maxim Rabinovich, Mitchell Stern, and Dan Klein. 2017.
\newblock \href {https://doi.org/10.18653/v1/P17-1105} {Abstract syntax networks for code generation and semantic parsing}.
\newblock In \emph{Proceedings of the 55th Annual Meeting of the Association for Computational Linguistics (Volume 1: Long Papers)}, pages 1139--1149, Vancouver, Canada. Association for Computational Linguistics.

\bibitem[{Semo et~al.(2022)Semo, Bernsohn, Hagag, Hayat, and Niklaus}]{CAP}
Gil Semo, Dor Bernsohn, Ben Hagag, Gila Hayat, and Joel Niklaus. 2022.
\newblock \href {https://doi.org/10.18653/v1/2022.nllp-1.3} {{C}lass{A}ction{P}rediction: A challenging benchmark for legal judgment prediction of class action cases in the {US}}.
\newblock In \emph{Proceedings of the Natural Legal Language Processing Workshop 2022}, pages 31--46, Abu Dhabi, United Arab Emirates (Hybrid). Association for Computational Linguistics.

\bibitem[{Suzgun et~al.(2023)Suzgun, Scales, Sch{\"a}rli, Gehrmann, Tay, Chung, Chowdhery, Le, Chi, Zhou, and Wei}]{suz2022}
Mirac Suzgun, Nathan Scales, Nathanael Sch{\"a}rli, Sebastian Gehrmann, Yi~Tay, Hyung~Won Chung, Aakanksha Chowdhery, Quoc Le, Ed~Chi, Denny Zhou, and Jason Wei. 2023.
\newblock \href {https://doi.org/10.18653/v1/2023.findings-acl.824} {Challenging {BIG}-bench tasks and whether chain-of-thought can solve them}.
\newblock In \emph{Findings of the Association for Computational Linguistics: ACL 2023}, pages 13003--13051, Toronto, Canada. Association for Computational Linguistics.

\bibitem[{Trivedi et~al.(2023)Trivedi, Balasubramanian, Khot, and Sabharwal}]{trivedi2023interleavingretrievalchainofthoughtreasoning}
Harsh Trivedi, Niranjan Balasubramanian, Tushar Khot, and Ashish Sabharwal. 2023.
\newblock \href {https://doi.org/10.18653/v1/2023.acl-long.557} {Interleaving retrieval with chain-of-thought reasoning for knowledge-intensive multi-step questions}.
\newblock In \emph{Proceedings of the 61st Annual Meeting of the Association for Computational Linguistics (Volume 1: Long Papers)}, pages 10014--10037, Toronto, Canada. Association for Computational Linguistics.

\bibitem[{Wang et~al.(2023)Wang, Wei, Schuurmans, Le, Chi, Narang, Chowdhery, and Zhou}]{wang2023selfconsistencyimproveschainthought}
Xuezhi Wang, Jason Wei, Dale Schuurmans, Quoc~V Le, Ed~H. Chi, Sharan Narang, Aakanksha Chowdhery, and Denny Zhou. 2023.
\newblock \href {https://openreview.net/forum?id=1PL1NIMMrw} {Self-consistency improves chain of thought reasoning in language models}.
\newblock In \emph{The Eleventh International Conference on Learning Representations}.

\bibitem[{Wang et~al.(2024)Wang, Liu, Lin, Li, Ma, and Liang}]{wang2024ratretrievalaugmentedthoughts}
Zihao Wang, Anji Liu, Haowei Lin, Jiaqi Li, Xiaojian Ma, and Yitao Liang. 2024.
\newblock \href {http://arxiv.org/abs/2403.05313} {{RAT}: {Retrieval Augmented Thoughts} elicit context-aware reasoning in long-horizon generation}.

\bibitem[{Wang et~al.(2025)Wang, Asai, Yu, Xu, Xie, Neubig, and Fried}]{wang-etal-2025-coderag}
Zora~Zhiruo Wang, Akari Asai, Xinyan~Velocity Yu, Frank~F. Xu, Yiqing Xie, Graham Neubig, and Daniel Fried. 2025.
\newblock \href {https://doi.org/10.18653/v1/2025.findings-naacl.176} {{C}ode{RAG}-bench: Can retrieval augment code generation?}
\newblock In \emph{Findings of the Association for Computational Linguistics: NAACL 2025}, pages 3199--3214, Albuquerque, New Mexico. Association for Computational Linguistics.

\bibitem[{Wei et~al.(2022)Wei, Wang, Schuurmans, Bosma, Ichter, Xia, Chi, Le, and Zhou}]{CoT}
Jason Wei, Xuezhi Wang, Dale Schuurmans, Maarten Bosma, Brian Ichter, Fei Xia, Ed~H. Chi, Quoc~V. Le, and Denny Zhou. 2022.
\newblock Chain-of-thought prompting elicits reasoning in {Large {Language Models}}.
\newblock In \emph{Proceedings of the 36th International Conference on Neural Information Processing Systems}, NIPS '22, Red Hook, NY, USA. Curran Associates Inc.

\bibitem[{Winter(2016)}]{Winter2016Elements}
Yoad Winter. 2016.
\newblock \href {https://books.google.com/books/about/Elements\_of\_Formal\_Semantics.html?hl\=\&id\=gioONQEACAAJ} {\emph{Elements of Formal Semantics}}.
\newblock Edinburgh Advanced Textbooks in Linguistics.

\bibitem[{Xia et~al.(2023)Xia, Wei, and Zhang}]{xia}
Chunqiu~Steven Xia, Yuxiang Wei, and Lingming Zhang. 2023.
\newblock \href {https://doi.org/10.1109/ICSE48619.2023.00129} {Automated program repair in the era of large pre-trained language models}.
\newblock In \emph{2023 IEEE/ACM 45th International Conference on Software Engineering (ICSE)}, pages 1482--1494.

\bibitem[{Xiao et~al.(2018)Xiao, Zhong, Guo, Tu, Liu, Sun, Feng, Han, Hu, Wang, and Xu}]{CAIL}
Chaojun Xiao, Haoxi Zhong, Zhipeng Guo, Cunchao Tu, Zhiyuan Liu, Maosong Sun, Yansong Feng, Xianpei Han, Zhen Hu, Heng Wang, and Jianfeng Xu. 2018.
\newblock \href {http://arxiv.org/abs/1807.02478} {{CAIL}2018: A large-scale legal dataset for judgment prediction}.

\bibitem[{Yang et~al.(2024)Yang, Chen, and Tam}]{yang2024prolog}
Xiaocheng Yang, Bingsen Chen, and Yik-Cheung Tam. 2024.
\newblock \href {https://aclanthology.org/2024.naacl-short.61/} {Arithmetic reasoning with {LLM}: {P}rolog generation \& permutation}.
\newblock In \emph{Proceedings of the 2024 Conference of the North American Chapter of the Association for Computational Linguistics: Human Language Technologies (Volume 2: Short Papers)}, pages 699--710. Association for Computational Linguistics.

\bibitem[{Yao et~al.(2023)Yao, Yu, Zhao, Shafran, Griffiths, Cao, and Narasimhan}]{10.5555/3666122.3666639}
Shunyu Yao, Dian Yu, Jeffrey Zhao, Izhak Shafran, Thomas~L. Griffiths, Yuan Cao, and Karthik Narasimhan. 2023.
\newblock Tree of thoughts: deliberate problem solving with {Large Language Models}.
\newblock In \emph{Proceedings of the 37th International Conference on Neural Information Processing Systems}, NIPS '23, Red Hook, NY, USA. Curran Associates Inc.

\bibitem[{Yin and Neubig(2017)}]{yin-neubig-2017-syntactic}
Pengcheng Yin and Graham Neubig. 2017.
\newblock \href {https://doi.org/10.18653/v1/P17-1041} {A syntactic neural model for general-purpose code generation}.
\newblock In \emph{Proceedings of the 55th Annual Meeting of the Association for Computational Linguistics (Volume 1: Long Papers)}, pages 440--450, Vancouver, Canada. Association for Computational Linguistics.

\bibitem[{Yu et~al.(2023)Yu, Jiang, Clark, and Sabharwal}]{ifqa}
Wenhao Yu, Meng Jiang, Peter Clark, and Ashish Sabharwal. 2023.
\newblock \href {https://doi.org/10.18653/v1/2023.emnlp-main.515} {{I}f{QA}: A dataset for open-domain question answering under counterfactual presuppositions}.
\newblock In \emph{Proceedings of the 2023 Conference on Empirical Methods in Natural Language Processing}, pages 8276--8288, Singapore. Association for Computational Linguistics.

\bibitem[{Zheng et~al.(2025)Zheng, Chen, Li, Li, Zong, Shi, Xu, Song, Wong, and See}]{zheng2025cursecotlimitationschainofthought}
Tianshi Zheng, Yixiang Chen, Chengxi Li, Chunyang Li, Qing Zong, Haochen Shi, Baixuan Xu, Yangqiu Song, Ginny Wong, and Simon See. 2025.
\newblock \href {https://openreview.net/forum?id=7SIrvcYNYj} {The curse of cot: On the limitations of chain-of-thought in in-context learning}.
\newblock \emph{Transactions on Machine Learning Research}.

\end{thebibliography}
\bibliographystyle{acl_natbib}
\fussy
\iftaclpubformat

\onecolumn
\fi
\clearpage

\appendix

\section{Prompt Templates}
\label{app:prompts}

\blue{This appendix reports the three prompt templates used in \name{}: the initial codification prompt, the iterative revision prompt, and the $R_2$ snippet-generation prompt.}

\subsection{Initial codification}
\begin{quote}
\blue{{You are given a benchmark instance described in natural language.}}

\blue{{Your task is to translate the input into an executable Python program.}}

\blue{{The program should be executable as a standalone solution for the given instance, make important conditions explicit in code, and solve the instance by executable procedure rather than free-form explanation.}}

\blue{{Important reasoning steps should be expressed through executable statements rather than comments alone.}}

\blue{{[Benchmark input $t$ inserted here.]}}

\blue{{Return only the Python program.}}
\end{quote}

\subsection{Iterative revision}
\begin{quote}
\blue{{You are revising an executable Python program for a benchmark instance.}}

\blue{{Given the benchmark input, the current program, and the retrieved evidence, produce (i) a revised Python program and (ii) a retrieval query for missing information, if needed.}}

\blue{{The revised program should preserve correct parts where possible and make supported conditions, procedures, exceptions, and constraints more explicit in code.}}

\blue{{[Benchmark input, current code, retrieved evidence, and optional retrieved snippets inserted here.]}}

\blue{{Output format: [REVISED\_PROGRAM] ... [RETRIEVAL\_QUERY] ...}}
\end{quote}

\subsection{$R_2$ snippet generation}
\begin{quote}
\blue{{You are given a solved benchmark example.}}

\blue{{Your task is to produce a short Python snippet that captures a useful condition, check, or procedure from the example in executable form.}}

\blue{{The snippet should provide structural guidance for solving similar benchmark instances.}}

\blue{{Requirements:
- Use Python.
- Keep the snippet short and focused.
- Make the relevant condition, check, or procedure explicit in code.
- Prefer reusable executable structure over benchmark-specific surface wording.
- Do not include free-form explanation outside the code.}}

\blue{{Return only the Python snippet.}}
\end{quote}

\section{Examples of Computables and Execution Traces}
\subsection{Proof in Uniform Topology}
\label{app:example_proofnet}

\blue{This case is not solved by textual baselines.}

\paragraph{Task overview.}
\blue{This \textsc{ProofNet} instance asks to show that}
\[
X = [0,1]^\omega
\]
\blue{is not locally compact under the uniform topology.}

\paragraph{Computable excerpt.}
\blue{The generated program constructs explicit points and checks the condition needed to show that the space is not locally compact.}

\begin{lstlisting}[style=smallpython,caption={Excerpt of a generated computable for the local non-compactness proof.}]
# witness in [0,r]^\omega
def y(k, r, N=20):
    v = [0.0] * N
    v[k] = r
    return v
def d(x, y):
    return max(abs(a - b) for a, b in zip(x, y))
r = 0.3
print(d(y(2, r), y(7, r)))   # = r
\end{lstlisting}

\paragraph{Interpretation of the code.}
\blue{The program makes the proof step explicit by constructing concrete points and checking their distance under the uniform metric.}
\blue{Rather than leaving the argument in free-form text, it turns the relevant proof step into an executable procedure.}

\subsection{Rule-Heavy Defeasible Reasoning}
\label{app:example_boardgameqa}

\blue{This case is not solved by textual baselines.}

\paragraph{Task overview.}
\blue{This \textsc{BBEH}/BoardgameQA instance asks whether a target relation, \texttt{shouts\_at(finch, mermaid)}, follows from a large collection of facts and defeasible rules.}
\blue{This instance requires a multi-step derivation involving explicit negation, chained rule application, and priority relations between competing conclusions.}

\blue{The generated program implements a defeasible inference procedure with signed literals, rule application, and priority-based conflict resolution.}

\begin{lstlisting}[style=smallpython,caption={Excerpt of the generated inference engine for a rule-heavy BoardgameQA instance.}]
@dataclass
class Rule:
    name: str
    condition: Callable[["Engine"], bool]
    conclusion: Callable[["Engine"], List[Literal]]
class Engine:
    def __init__(self):
        self.supports: Dict[Literal, Set[str]] = {}
        self.rules: List[Rule] = []
        self.preferences: Set[Tuple[str, str]] = set()
    def add_derived(self, x: Literal, rule_name: str) -> bool:
        ox = opposite(x)
        if ox in self.supports:
            opp_supports = self.supports[ox]
            if self.stronger_than_all(rule_name, opp_supports):
                del self.supports[ox]
            else:
                return False
        self.supports[x] = {rule_name}
        return True
    def run(self):
        changed = True
        while changed:
            changed = False
            for rule in self.rules:
                if rule.condition(self):
                    for c in rule.conclusion(self):
                        if self.add_derived(c, rule.name):
                            changed = True
\end{lstlisting}

\paragraph{Query and execution outcome.}
\blue{The program evaluates the target query directly:}

\begin{lstlisting}[style=smallpython,caption={Final query evaluation in the generated program.}]
query = lit("shouts_at", "finch", "mermaid")
answer = E.explain_query(query)
print("Answer:", answer)
\end{lstlisting}

\paragraph{Derivation summary.}
\blue{The final judgment is obtained through a multi-step chain of rule applications that derives the negated target relation.}

\subsection{Example of Silent Failure}
\label{app:example_silent_failure}

\blue{This example illustrates a case in which the generated program executes successfully but still produces an incorrect result for the original benchmark instance.}

\paragraph{Task overview.}
\blue{This \textsc{BBEH} hyperbaton instance asks which option uses the correct adjective order in a variant of English. The gold answer is \textbf{J}.}

\paragraph{Computable excerpt.}
\blue{The generated program infers an adjective-class order from the training examples, assigns each adjective in the candidate options to a class, and then checks whether the resulting class sequence is sorted according to the inferred order. The relevant excerpt is shown below.}

\begin{lstlisting}[style=smallpython,caption={Excerpt of a generated program that executes successfully but returns an incorrect answer.}]
def solve_adjective_order():
    # Inferred adjective class order from the examples:
    # origin > shape > size > purpose > material > age > color > opinion
    order = {
        "origin": 0,
        "shape": 1,
        "size": 2,
        "purpose": 3,
        "material": 4,
        "age": 5,
        "color": 6,
        "opinion": 7,
    }
    def is_correct(sentence: str) -> bool:
        words = sentence.split()
        classes = []
        for w in words[:-1]:
            if w in adjective_class:
                classes.append(adjective_class[w])
            else:
                return False
        ranks = [order[c] for c in classes]
        return ranks == sorted(ranks)

    correct = []
    for key, sent in options.items():
        if is_correct(sent):
            correct.append(key)
    final_answer = "".join(correct) if correct else "K"
    print("Final answer:", final_answer)
\end{lstlisting}

\paragraph{Execution outcome.}
\blue{The program executes without runtime error and outputs \textbf{ACDEGJ}. The gold answer, however, is \textbf{J}.}

\paragraph{Why this is a silent failure.}
\blue{This is a silent failure because the program runs successfully but still gives the wrong result for the original benchmark instance. The failure is therefore not caused by syntax, non-executability, or runtime exception. Instead, the induced procedure itself is semantically incorrect for the task: it imposes an inferred global ordering that does not correctly capture the adjective-order system required by the instance. This example shows that successful execution does not by itself guarantee semantic correctness.}
\begin{table*}[t]
\centering
\caption{Additional-model evaluation across all datasets: Claude and Gemini.}
\vspace{-3mm}
\label{tab:additional_models_a}
\footnotesize
\setlength{\tabcolsep}{3.6pt}
\renewcommand{\arraystretch}{0.95}
\begin{tabular}{l|cccccc|cccccc}
\toprule
& \multicolumn{6}{c|}{\textbf{Claude}} 
& \multicolumn{6}{c}{\textbf{Gemini}} \\
\textbf{Dataset}
& \textbf{CoT} & \textbf{CoC} & \textbf{Ours$_{\textsc{NL}}$} & \textbf{Ours} & \textbf{Exec (\%)} & \makecell[c]{\textbf{$N$}\\\textbf{(Avg/Max)}}
& \textbf{CoT} & \textbf{CoC} & \textbf{Ours$_{\textsc{NL}}$} & \textbf{Ours} & \textbf{Exec (\%)} & \makecell[c]{\textbf{$N$}\\\textbf{(Avg/Max)}} \\
\midrule
GSM8K       & 0.95 & 0.93 & 0.97 & 0.98 & 99 & 1.18/3  & 0.97 & 0.98 & 0.97 & 0.98 & 99 & 1.10/2 \\
ProofNet    & 0.86 & 0.88 & 0.90 & 0.92 & 96 & 2.05/6  & 0.88 & 0.91 & 0.93 & 0.94 & 98 & 1.78/5 \\
IFQA        & 0.16 & 0.28 & 0.34 & 0.39 & 91 & 1.48/4  & 0.32 & 0.36 & 0.52 & 0.52 & 96 & 1.31/3 \\
Corr2Cause  & 0.77 & 0.79 & 0.79 & 0.81 & 93 & 2.26/8  & 0.84 & 0.81 & 0.86 & 0.88 & 97 & 2.05/7 \\
BBH         & 0.58 & 0.62 & 0.64 & 0.64 & 88 & 1.36/4  & 0.75 & 0.71 & 0.81 & 0.89 & 91 & 1.18/3 \\
BBEH        & 0.11 & 0.27 & 0.30 & 0.35 & 86 & 1.52/5  & 0.16 & 0.22 & 0.28 & 0.36 & 87 & 1.30/4 \\
CAIL        & 0.42 & 0.47 & 0.63 & 0.68 & 93 & 2.11/6  & 0.55 & 0.59 & 0.73 & 0.77 & 95 & 1.88/5 \\
ECHR        & 0.39 & 0.41 & 0.52 & 0.55 & 86 & 2.84/10 & 0.43 & 0.49 & 0.63 & 0.79 & 94 & 2.32/8 \\
CAP         & 0.58 & 0.57 & 0.63 & 0.65 & 94 & 2.47/9  & 0.61 & 0.63 & 0.66 & 0.67 & 92 & 2.18/8 \\
PubMedQA    & 0.59 & 0.66 & 0.73 & 0.75 & 90 & 1.86/5  & 0.72 & 0.66 & 0.84 & 0.86 & 95 & 1.66/4 \\
USMLE       & 0.69 & 0.75 & 0.74 & 0.76 & 91 & 2.31/8  & 0.73 & 0.81 & 0.84 & 0.87 & 97 & 2.11/7 \\
HealthClaim & 0.37 & 0.42 & 0.44 & 0.46 & 92 & 1.21/3  & 0.55 & 0.59 & 0.69 & 0.72 & 96 & 1.08/2 \\
BioASQ-6B   & 0.75 & 0.82 & 0.83 & 0.85 & 94 & 1.34/4  & 0.76 & 0.83 & 0.89 & 0.93 & 96 & 1.18/3 \\
\bottomrule
\end{tabular}
\vspace{-3mm}
\end{table*}

\section{Additional Experimental Settings}

\label{app:repro}
\paragraph{Code memory construction.}
\blue{ As described in
  the main text (Section 4.3), our $R_2$ repository is leakage-safe;
  that is, each task is solved without using its own dataset. This is
  ensured by using cross-validation to exclude the dataset chunk that
  contains the task.}
\blue{
  For each validation round, the $R_2$ repository is constructed
  offline in two steps. First, each source passage is translated into
  a single Python snippet using Prompt~A.3
  (\autoref{app:prompts}). Second, the resulting index--knowledge
  pairs are stored in a {\textsc{FAISS} index \cite{8733051}}
  without any additional ontology or schema. This repository stores
  short executable snippets that are retrieved during iterative
  revision as structural guidance.}

\paragraph{Runtime and retrieval settings.}
\blue{We use the chat-completion interface with a single system message and model temperature $\tau_{\text{code}}=0.2$.}
\blue{The retrieval pipeline uses \textsc{all-MiniLM-L6-v2} embeddings (dimension 384) and returns the top-3 nearest code snippets from the offline index.}

\paragraph{Scope of release.}
\blue{We release representative computables and the prompt-level
  protocol needed to inspect the workflow.}  \blue{More complete
  implementation materials, including internal scripts, are not
  released at this stage because the associated system is currently
  under review.}  \blue{We plan to release these
  materials after the review process is
  completed.}

\section{Additional Models}
\label{app:additional_models}

\blue{Table~\ref{tab:additional_models_a} and Table~\ref{tab:additional_models_b} show that the main pattern of the paper transfers beyond the primary GPT-5-based setting.}
\blue{We evaluate \name{} on Claude, Gemini, Qwen, and Llama.}
\blue{The specific model versions used in these experiments were \texttt{claude-haiku-4-5-202510}, \texttt{gemini-3.1-pro-preview}, Qwen2.5-72B-Instruct, and Llama-3.3-70B-Instruct.}

\blue{For these additional experiments, we also report the average number of refinement rounds, denoted by \(N\), in the same format as in the main results table.}
\blue{This is included to show whether the iterative behavior of the system remains comparable across models as well as across datasets.}

\begin{table*}[t]
\centering
\caption{Additional-model evaluation across all datasets: Qwen and Llama.}
\vspace{-3mm}
\label{tab:additional_models_b}
\footnotesize
\setlength{\tabcolsep}{3.6pt}
\renewcommand{\arraystretch}{0.95}
\begin{tabular}{l|cccccc|cccccc}
\toprule
& \multicolumn{6}{c|}{\textbf{Qwen}} 
& \multicolumn{6}{c}{\textbf{Llama}} \\
\textbf{Dataset}
& \textbf{CoT} & \textbf{CoC} & \textbf{Ours$_{\textsc{NL}}$} & \textbf{Ours} & \textbf{Exec (\%)} & \makecell[c]{\textbf{$N$}\\\textbf{(Avg/Max)}}
& \textbf{CoT} & \textbf{CoC} & \textbf{Ours$_{\textsc{NL}}$} & \textbf{Ours} & \textbf{Exec (\%)} & \makecell[c]{\textbf{$N$}\\\textbf{(Avg/Max)}} \\
\midrule
GSM8K       & 0.94 & 0.95 & 0.96 & 0.96 & 99 & 1.24/3  & 0.95 & 0.96 & 0.96 & 0.97 & 100 & 1.20/3 \\
ProofNet    & 0.82 & 0.86 & 0.88 & 0.88 & 89 & 2.12/6  & 0.80 & 0.84 & 0.88 & 0.93 & 91  & 2.04/6 \\
IFQA        & 0.21 & 0.27 & 0.33 & 0.36 & 91 & 1.56/4  & 0.30 & 0.28 & 0.46 & 0.43 & 99  & 1.44/4 \\
Corr2Cause  & 0.64 & 0.61 & 0.67 & 0.71 & 91 & 2.34/8  & 0.53 & 0.54 & 0.71 & 0.73 & 95  & 2.29/8 \\
BBH         & 0.52 & 0.58 & 0.66 & 0.68 & 92 & 1.42/4  & 0.57 & 0.56 & 0.68 & 0.70 & 99  & 1.39/4 \\
BBEH        & 0.12 & 0.19 & 0.24 & 0.25 & 84 & 1.61/5  & 0.14 & 0.18 & 0.23 & 0.27 & 82  & 1.57/5 \\
CAIL        & 0.59 & 0.64 & 0.69 & 0.73 & 93 & 1.96/5  & 0.52 & 0.61 & 0.72 & 0.76 & 100 & 2.18/6 \\
ECHR        & 0.16 & 0.23 & 0.39 & 0.43 & 87 & 2.96/10 & 0.27 & 0.31 & 0.49 & 0.53 & 97  & 2.71/9 \\
CAP         & 0.47 & 0.49 & 0.54 & 0.58 & 90 & 2.63/9  & 0.46 & 0.50 & 0.56 & 0.60 & 85  & 2.41/8 \\
PubMedQA    & 0.66 & 0.64 & 0.74 & 0.76 & 91 & 1.94/5  & 0.70 & 0.68 & 0.76 & 0.77 & 94  & 1.82/5 \\
USMLE       & 0.57 & 0.59 & 0.67 & 0.67 & 93 & 2.52/8  & 0.71 & 0.73 & 0.77 & 0.77 & 93  & 2.20/7 \\
HealthClaim & 0.30 & 0.38 & 0.46 & 0.47 & 94 & 1.34/3  & 0.51 & 0.53 & 0.57 & 0.58 & 92  & 1.16/3 \\
BioASQ-6B   & 0.82 & 0.84 & 0.86 & 0.85 & 97 & 1.41/4  & 0.77 & 0.75 & 0.80 & 0.79 & 94  & 1.26/3 \\
\bottomrule
\end{tabular}
\vspace{-5mm}
\end{table*}

\blue{Across Claude, Gemini, Qwen, and Llama, \textsc{Ours} remains the strongest method overall and outperforms both \textsc{CoT} and \textsc{CoC} across benchmark families.}
\blue{The gains are largest on causal, legal, and biomedical benchmarks, where correct decisions depend on recovering background rules, exceptions, or domain knowledge and turning them into explicit checks.}

\blue{At the same time, the table suggests that executable codification alone is less stable across models than the full iterative framework.
  The jump from \textsc{CoT} to \textsc{CoC} is often modest or inconsistent, especially for some open-weight models, whereas the gains from \textsc{Ours}$_{\textsc{NL}}$
  {(our proposed method without code snippets in RAG)}
   and \textsc{Ours} are much more systematic.
This indicates that the cross-model gains are not explained by one-shot code generation alone, but also depend on iterative retrieval-guided restructuring.
The full method remains strongest when this restructuring is combined with executable codification and execution-based revision.}

\blue{Finally, executability remains high across models, indicating that the workflow itself transfers beyond the primary model setting even when absolute accuracy differs by model.
At the same time, the lower scores of some open-weight models show that the framework does not remove underlying model differences: it improves weaker models, but does not fully close the gap to the strongest systems.}

\section{Lean Feasibility Probe}
\label{app:lean}

\blue{This appendix gives one compact example of the hand-templated Lean setup used in the feasibility probe discussed in Section~7.1. The goal of this comparison is not to provide a full formal-semantics baseline, but to test whether benchmark instances can be instantiated in a much stronger proof-oriented environment.}

\paragraph{Example: CAIL accusation template.}
\blue{For the CAIL accusation task, the Lean artifact is generated under a fixed interface. The template requires the generated file to: (i) use only \texttt{import Std}; (ii) define \texttt{predict (fact : String) : String}; (iii) define \texttt{feats (fact : String) : List (String × Bool)}; (iv) read case facts from standard input; and (v) restrict the output label to a predefined accusation vocabulary or \texttt{Unknown}.}
\blue{This setup illustrates the role of hand-templated encoding in the Lean probe: the overall interface, admissible labels, and input/output structure are fixed in advance, while the model is responsible for producing an instance-specific Lean artifact.}

\paragraph{Observed failure modes.}
\blue{Across tasks, we observed three main failure modes: some instances require assumptions or intermediate lemmas that are difficult to encode within a fixed template; some benchmark descriptions are too open-textured to map cleanly onto a predetermined formal skeleton; and some generated scripts fail because the model does not reliably produce well-typed Lean code.}

\paragraph{Scope of the comparison.}
\blue{The Lean comparison is therefore best understood as a feasibility probe. Its role in the paper is to illustrate the practical gap between proof-oriented formalization and the weaker executable representations studied in the main framework.}
\end{document}